\documentclass[final,3p,times]{elsarticle}
\usepackage{lineno}
\usepackage[hidelinks]{hyperref}
\usepackage{graphicx}
\usepackage{amsmath,amssymb} 
\usepackage{booktabs,multirow,adjustbox,makecell}

\usepackage{caption}
\usepackage{tabularx}
\usepackage{subcaption}
\modulolinenumbers[5]

\begin{document}
\begin{frontmatter}

\title{Hybrid Deep Learning with Temporal Data Augmentation for Accurate Remaining Useful Life Prediction of Lithium-Ion Batteries}

\author[1,*]{Yun Tian}
\author[1,*]{Guili Wang}
\author[2,*]{Jian Bi}
\author[3]{Kaixin Han}
\author[1]{Chenglu Wu}
\author[1]{Zhiyi Lu}
\author[1]{Chenhao Li}
\author[1]{Liangwang Sun}
\author[1]{Minyu Zhou}
\author[1,4,5]{\texorpdfstring{Chenchen Xu$^{\dagger}$}{Chenchen Xu}}

% 单位地址
\address[1]{School of Physics and Electronic Information, Anhui Normal University, Wuhu 241002, China}
\address[2]{College of Mechanical and Vehicle Engineering, Hunan University, Changsha 410082, China}
\address[3]{School of Design, NingboTech University, Ningbo 315100, China}
\address[4]{The Chinese University of Hong Kong, Shatin, Hong Kong SAR, China}
\address[5]{Shenzhen Bay Laboratory, Shenzhen 518132, China}

% 通讯作者脚注
\cortext[corr]{Corresponding author. E-mail address: xuchenchen@zju.edu.cn (C. Xu).}

% 共同第一作者脚注  
\fntext[*]{These authors contributed equally to this work.}

\begin{abstract}
Accurate prediction of lithium-ion battery remaining useful life (RUL) is essential for reliable health monitoring and data-driven analysis of battery degradation. However, the robustness and generalization capabilities of existing RUL prediction models are significantly challenged by complex operating conditions and limited data availability. To address these limitations, this study proposes a hybrid deep learning model, CDFormer, which integrates convolutional neural networks, deep residual shrinkage networks, and Transformer encoders extract multiscale temporal features from battery measurement signals, including voltage, current, and capacity. This architecture enables the joint modeling of local and global degradation dynamics, effectively improving the accuracy of RUL prediction.To enhance predictive reliability, a composite temporal data augmentation strategy is proposed, incorporating Gaussian noise, time warping, and time resampling, explicitly accounting for measurement noise and variability. CDFormer is evaluated on two real-world datasets, with experimental results demonstrating its consistent superiority over conventional recurrent neural network-based and Transformer-based baselines across key metrics. By improving the reliability and predictive performance of RUL prediction from measurement data, CDFormer provides accurate and reliable forecasts, supporting effective battery health monitoring and data-driven maintenance strategies.
\end{abstract}

\end{frontmatter}

\section{Introduction}
In recent years, lithium-ion batteries have become critical energy storage devices in both industrial and everyday applications due to their high energy density, long cycle life, and decreasing cost~\cite{hasan2025advancing, sharma2025critical}. In this context, accurate monitoring and prediction of battery degradation are essential for ensuring safe and reliable operation in practical applications. However, prolonged cycling leads to capacity fade and increased internal resistance, resulting in performance degradation~\cite{ibraheem2023capacity, li2024evolution}. Therefore, accurately predicting the remaining useful life (RUL) of batteries from measurement data is particularly important for effective health monitoring and maintenance management~\cite{ahwiadi2025battery, waseem2025advancement}.

Traditional RUL prediction increasingly relies on deep learning techniques because they can model complex temporal dependencies in battery degradation data~\cite{zhaoa2021review}. Recurrent neural networks (RNNs) and their variants, such as long short-term memory (LSTM) networks~\cite{ZhaoLSTM2022} and gated recurrent units (GRUs)\cite{Lin2023}, have been widely used to capture sequential features\cite{Catelani2021}. However, RNNs can suffer from gradient vanishing or exploding, which makes training on long sequences difficult. LSTMs address this issue by introducing gating mechanisms that help maintain long-term memory. This change makes the model more complex and can lead to overfitting, especially when training data is limited. GRUs provide a more streamlined architecture with fewer parameters. Nevertheless, their ability to capture multi-scale temporal patterns is limited, which is essential for accurate battery degradation modeling.

recently, attention-based models like DeTransformer~\cite{Chen2022} have been proposed to overcome the limitations of RNN-based architectures by effectively capturing long-range dependencies. DeTransformer improves global sequence modeling, but it may have difficulty capturing detailed local features. These local features are important for describing complex degradation patterns. SA-LSTM~\cite{WangLSTM2023} uses self-attention to solve this problem. However, it increases model complexity and makes tuning harder. Furthermore, mixture-of-experts models such as AttMoE~\cite{ChenAttMoE2024} enhance adaptability and robustness by leveraging multiple specialized subnetworks, but they introduce additional architectural complexity and require extensive hyperparameter tuning, which may limit their practical application.

To address these challenges, we propose CDFormer, a hybrid deep learning framework that integrates convolutional neural networks (CNNs), deep residual shrinkage networks (DRSNs), and Transformer encoders. This architecture combines local feature extraction, noise-resilient representation, and long-range temporal modeling to improve the model’s ability to characterize battery degradation patterns.

Furthermore, to realistically simulate diverse battery degradation behaviors and complex aging patterns in practical and industrial scenarios, we implement three temporal data augmentation strategies: Gaussian noise injection, time warping, and time resampling. These strategies effectively emulate operational variations, including sensor noise, irregular charge–discharge cycles, and non-uniform sampling, thereby enhancing model robustness and generalization across heterogeneous real-world conditions.

Through both quantitative and qualitative evaluation on two real-world datasets, we demonstrate that the proposed method significantly outperforms conventional baselines and other deep learning-based variants. Compared with the state-of-the-art AttMoE~\cite{ChenAttMoE2024} method, our model shows remarkable improvements, achieving average error reductions of 24.6\% in root mean square error (RMSE), 30.4\% in mean absolute error (MAE), and 25.9\% in relative error (RE). These results highlight the model’s superior performance in battery lifespan prediction and demonstrate its strong potential for practical deployment.
The main contributions of this work are summarized as follows:
\begin{itemize}
    \item We develop CDFormer, a hybrid architecture that synergistically combines CNNs, DRSNs, and Transformers to improve robustness and accuracy in lithium-ion battery RUL prediction.
    \item We propose and integrate three temporal data augmentation methods, including two novel techniques—time warping and time resampling—which simulate diverse battery aging and operational patterns, thereby enhancing model generalization under complex real-world conditions.
    \item Comprehensive experiments on two real-world datasets demonstrate our method's superior performance over existing baselines and other deep learning-based variants.
\end{itemize}

\section{Related Work}
Existing RUL prediction methods can be broadly categorized into two groups: physics-based modeling approaches~\cite{el2023physics,BrosaPlanella2022,liu2023} and data-driven methods~\cite{Liu2020,Wu2016,Gou2020}. The former relies on a deep understanding of internal battery mechanisms, offering strong interpretability, while the latter leverages machine learning and deep learning techniques to automatically extract degradation patterns from large-scale historical monitoring data, demonstrating superior predictive performance. This section systematically reviews the development and representative methods of these two approaches to provide theoretical support and technical background for the subsequent model design.
\subsection{Physics-Based Modeling Methods}
Physics-based approaches have long been foundational in battery RUL prediction research, focusing on establishing models that relate battery aging processes to performance degradation to infer RUL. These methods primarily include empirical models~\cite{Lin2021,Stroe2014}, semiempirical models~\cite{DiFilippi2010,Prasad2013}, and electrochemical mechanism models~\cite{sadabadi2021prediction,Ramadass2004}. Empirical and semi-empirical models typically assume that battery capacity degradation follows certain mathematical expressions, such as exponential, polynomial, or hyperbolic functions, fitting the degradation trajectory accordingly to build life prediction models. For instance, Khelif et al.~\cite{khelif2015experience} proposed an RUL prediction method based on health indicators combined with regression models, while Xu et al.~\cite{Xu2016} developed a capacity degradation model incorporating stress factors such as the average state of charge (SOC) and the depth of discharge (DOD), grounded in the formation mechanism of the solid electrolyte interphase (SEI). 

In contrast, electrochemical modeling focuses on characterizing performance changes under multi-physical fields from the battery’s internal reactions. Uddin et al.~\cite{Uddin2016}, Eddahech et al.~\cite{Eddahech2012}, and Chang et al.~\cite{chang2022improvement} utilized electrochemical impedance spectroscopy (EIS) to model battery responses at different frequencies, analyzing interface characteristics to assess health status.
 Li et al.~\cite{Li2019} analyzed feature peaks in incremental capacity (IC) curves during charge-discharge cycles to infer aging levels and remaining life. differential voltage (DV)~\cite{Liu2015} and open circuit voltage (OCV)~\cite{Wang2022, fan2025state} analysis methods are also widely applied to model internal electrochemical changes for RUL estimation. Despite their theoretical interpretability and high accuracy under controlled experimental conditions, physics-based models face challenges in practice due to their heavy dependence on precise internal battery parameters, complex model construction, and limited generalizability. Furthermore, real-world operating conditions present inherent complexity and variability, characterized by highly dynamic environments and fluctuating load profiles. These factors introduce significant interference sources that existing models frequently fail to capture adequately, constraining their practical applicability in large-scale deployments.

\subsection{Deep Learning-Based Methods}
The widespread deployment of lithium-ion batteries not only advances energy infrastructure~\cite{tasneem2025lithium}, but also provides rich data and diverse real-world scenarios for deep learning-based remaining useful life (RUL) prediction. Various deep learning models~\cite{Ren2018,Wei2021,Zhao2023} have been widely applied to battery RUL prediction and achieved significant performance improvements on multiple public datasets~\cite{NASA2007,CALCE2022}. The RNN, as one of the earliest models applied to this task, utilizes its recurrent structure to capture temporal dependencies in sequential data but suffers from gradient vanishing issues, resulting in unstable modeling of long-term dependencies~\cite{wu2024remaining,Ansari2021}. To address this, the LSTM network, featuring forget and memory gates, has been successfully applied to battery health evaluation, effectively enhancing modeling of long-term degradation trends~\cite{Ren2020,ZhaoLSTM2022}. The GRU, a streamlined variant of LSTM, retains key gating mechanisms while offering improved training efficiency and robustness, becoming one of the mainstream choices for battery life prediction~\cite{Lin2023,wang2024lithium}. Beyond traditional recurrent structures, recent research has incorporated advanced attention mechanisms into RUL prediction models. The DeTransformer model introduces Transformer encoder architectures and self-attention mechanisms to extract long-range dependencies from capacity sequences, demonstrating stability in modeling nonlinear degradation trends~\cite{Chen2022}. The SA-LSTM model integrates attention mechanisms within the LSTM framework to enhance sensitivity to critical temporal information, improving prediction accuracy while maintaining structural stability~\cite{WangLSTM2023}. The AttMoE model combines mixture-of-experts (MoE) with attention mechanisms to dynamically weight different feature channels, exhibiting stronger adaptability to diverse aging patterns~\cite{ChenAttMoE2024}.

The above methods, each with distinct architectural designs, represent an evolutionary path from conventional recurrent neural networks to deep temporal modeling approaches fused with attention mechanisms. While these methods have improved battery RUL prediction, several limitations remain in practical energy storage applications: (1) Limited capability in modeling multi-scale temporal features: most models struggle to capture both short-term fluctuations and long-term degradation patterns, which are common across various battery applications;  (2) Dependence on single-dimensional input features: existing methods often use only single-variable time series, such as capacity or voltage, and fail to fully exploit multi-sensor information, including temperature, current, and environmental conditions;  (3) Poor generalization under noisy and dynamic conditions: real-world environments are complex and variable, with fluctuating loads, irregular usage patterns, and external disturbances, challenging the robustness of existing models.  

Therefore, developing a unified model that integrates multi-feature fusion, multi-scale temporal modeling, and robust noise resilience is a key challenge for improving the accuracy and reliability of battery life prediction. This challenge represents the core motivation and innovation of the present study.

\section{Method}
\begin{figure}[htbp]
    \centering
    \includegraphics[width=1\linewidth]{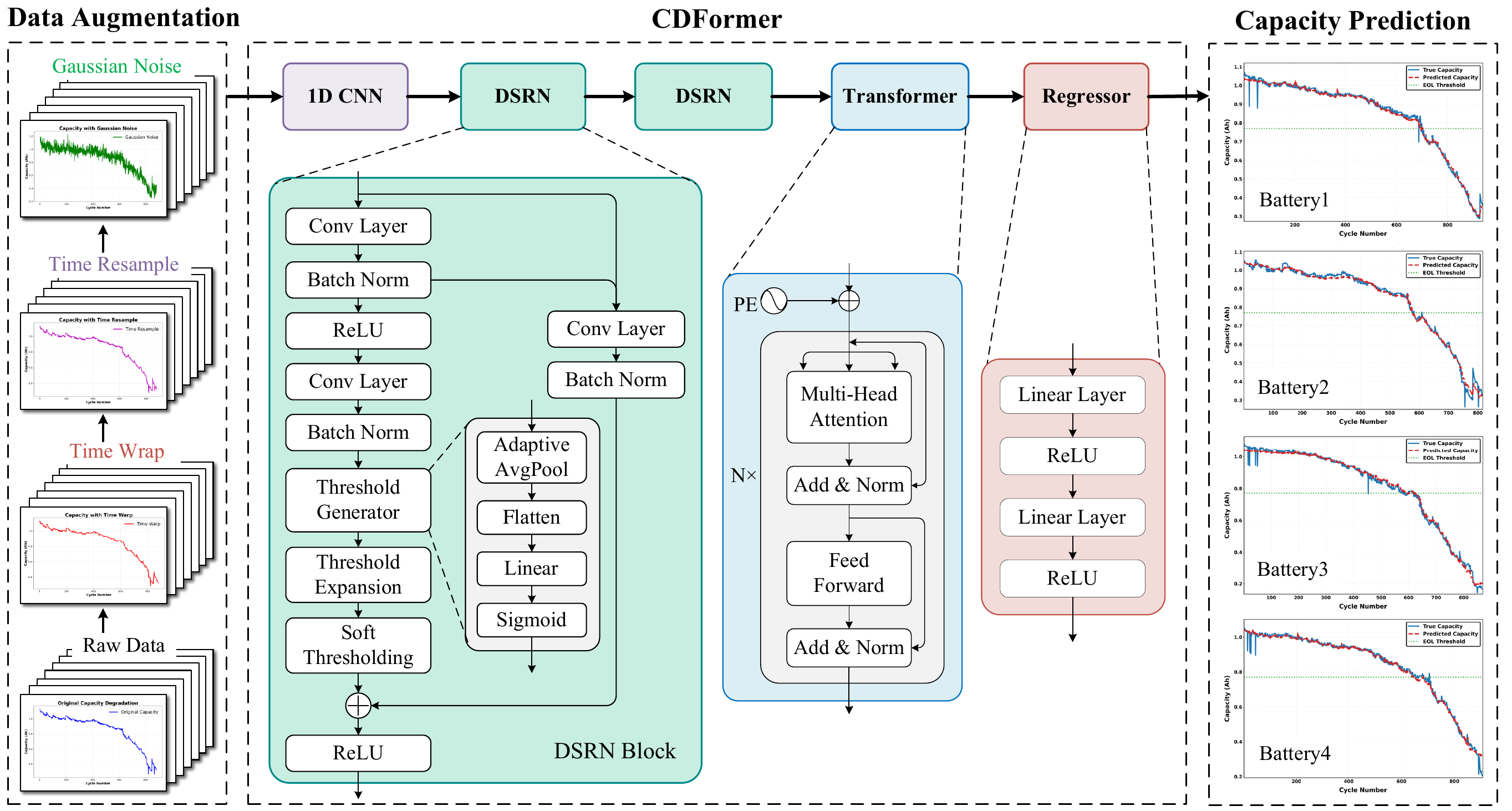}
    \caption{Workflow of the CDFormer prediction pipeline, including temporal data augmentation, feature extraction with CNN and DRSN, temporal dependency modeling via Transformer encoders, and RUL estimation.}
    \label{fig:CDFormer}
\end{figure}

In this study, we design a hybrid neural network model for predicting the RUL of lithium-ion batteries, grounded in the understanding of battery aging mechanisms and end-of-life (EOL) characteristics~\cite{madani2025comprehensive,Vetter2005,Guo2021EV}. The proposed model, named CDFormer, integrates CNN, DRSN, and Transformer encoders, aiming to fully leverage the strengths of each architecture in feature modeling. Specifically, CNNs~\cite{Koprinska2018,Ma2024} are utilized to efficiently extract local temporal features, DRSN~\cite{ZhaoShrinkage2019,WuDRSN2023} modules enhance feature representation through residual connections and soft-thresholding mechanisms, while Transformer~\cite{Vaswani2017,ZhouInformer2021} encoders are employed to capture long-range dependencies in sequential data. During training, we employ not only conventional Gaussian noise augmentation but also two temporal data augmentation techniques, namely time warping and time resampling, to enhance the model’s generalization ability and robustness against various degradation patterns. These augmentation techniques effectively expand the distributional diversity of the training data, enabling the model to learn richer temporal representations and perform more reliably under complex and variable real-world operating conditions. The Proposed method overview is illustrated in Fig.~\ref{fig:CDFormer}. The following subsections provide a detailed introduction to the CDFormer architecture and the implementation of the three temporal data augmentation techniques.

\subsection{CDFormer}
To capture the temporal degradation patterns of lithium-ion batteries, we propose CDFormer, a hybrid neural network. It combines one-dimensional CNNs (1D-CNNs), deep residual shrinkage networks (DRSNs), and Transformer encoders. The 1D-CNNs extract local temporal features. DRSNs enhance feature representation by suppressing noise and applying residual learning. Transformer encoders capture long-range dependencies. This design allows CDFormer to model multi-scale temporal information in battery degradation sequences effectively.

\subsubsection{1D-CNNs}
\begin{figure}[htbp]
    \centering
    \includegraphics[width=1\linewidth]{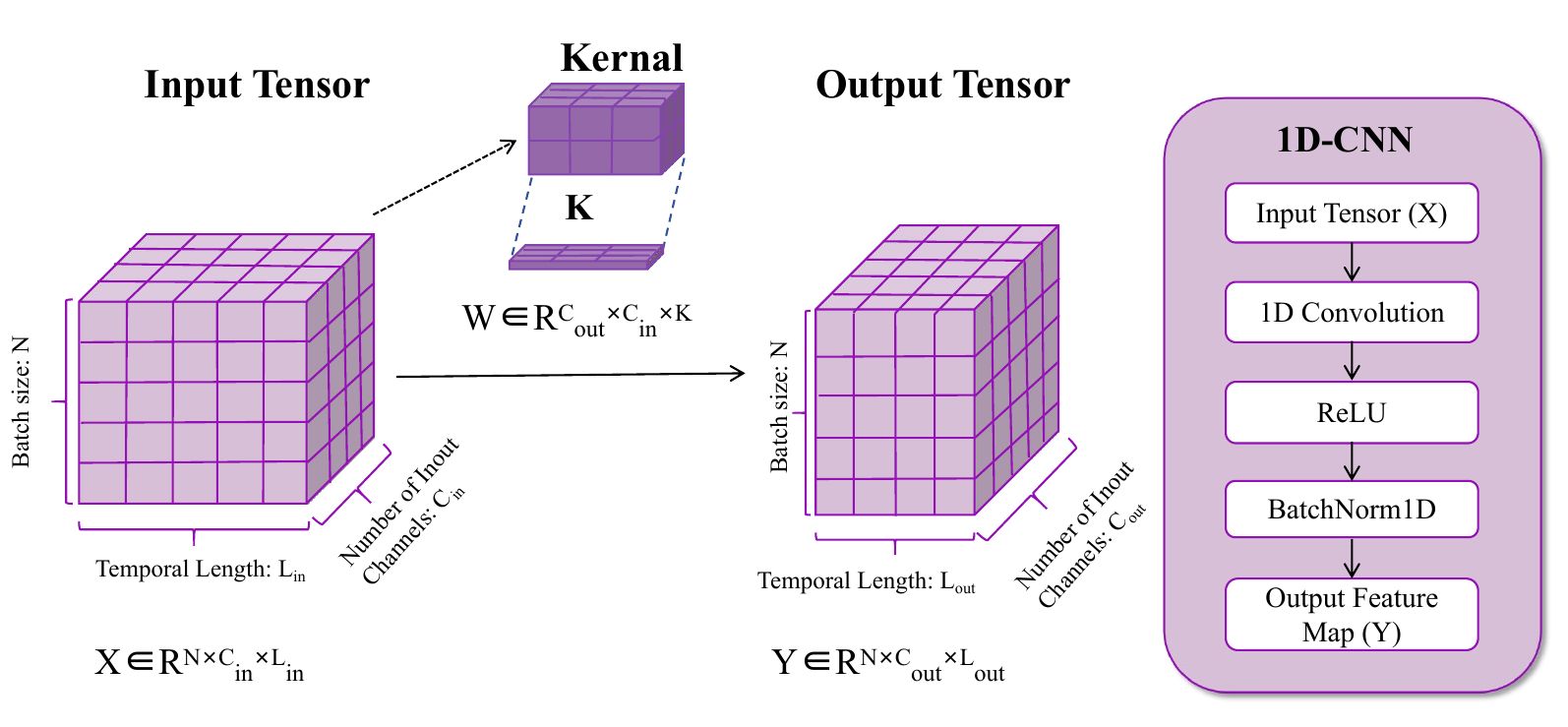}
    \caption{Overall architecture of the 1D-CNN model used for time-series feature extraction.}
    \label{fig:cnn1d_architecture}
\end{figure}

Known for their ability to extract local features, 1D-CNNs are particularly effective at capturing short-term dependencies and localized patterns in time-series data. Based on this advantage, we designed a 1D-CNN architecture illustrated in Fig.~\ref{fig:cnn1d_architecture} to effectively extract temporal features from the input data.

Let the input time-series be represented as a 3D tensor: \(X \in \mathbb{R}^{N \times C_{in} \times L_{in}}\), where \(N\) denotes the batch size, \(C_{in}\) is the number of input channels, and \(L_{in}\) represents the temporal length.  
The convolution kernel tensor is defined as: \(W \in \mathbb{R}^{C_{out} \times C_{in} \times K}\), where \(C_{out}\) indicates the number of output channels, and \(K\) specifies the kernel width.  
Each kernel slides along the temporal axis, performing weighted summation within local receptive fields to extract local patterns from the time-series. The corresponding output feature tensor \(Y \in \mathbb{R}^{N \times C_{\text{out}} \times L_{\text{out}}}\) is computed as follows:
\begin{equation}
Y[n, c_{\text{out}}, t] = \sum_{c_{\text{in}}=1}^{C_{\text{in}}} \sum_{k=1}^{K} X[n, c_{\text{in}}, t + k] \cdot W[c_{\text{out}}, c_{\text{in}}, k] + b[c_{\text{out}}],
\label{eq:cnn1d}
\end{equation}
where \(b \in \mathbb{R}^{C_{\text{out}}}\) denotes the bias term. This module autonomously learns discriminative features from input data, significantly reducing manual intervention in feature engineering. Furthermore, 1D-CNNs preserve temporal ordering information, making them particularly suitable for tasks involving time-dependent patterns.

\subsubsection{DRSN}
\begin{figure}[htbp]
\centering
\includegraphics[width=1\textwidth]{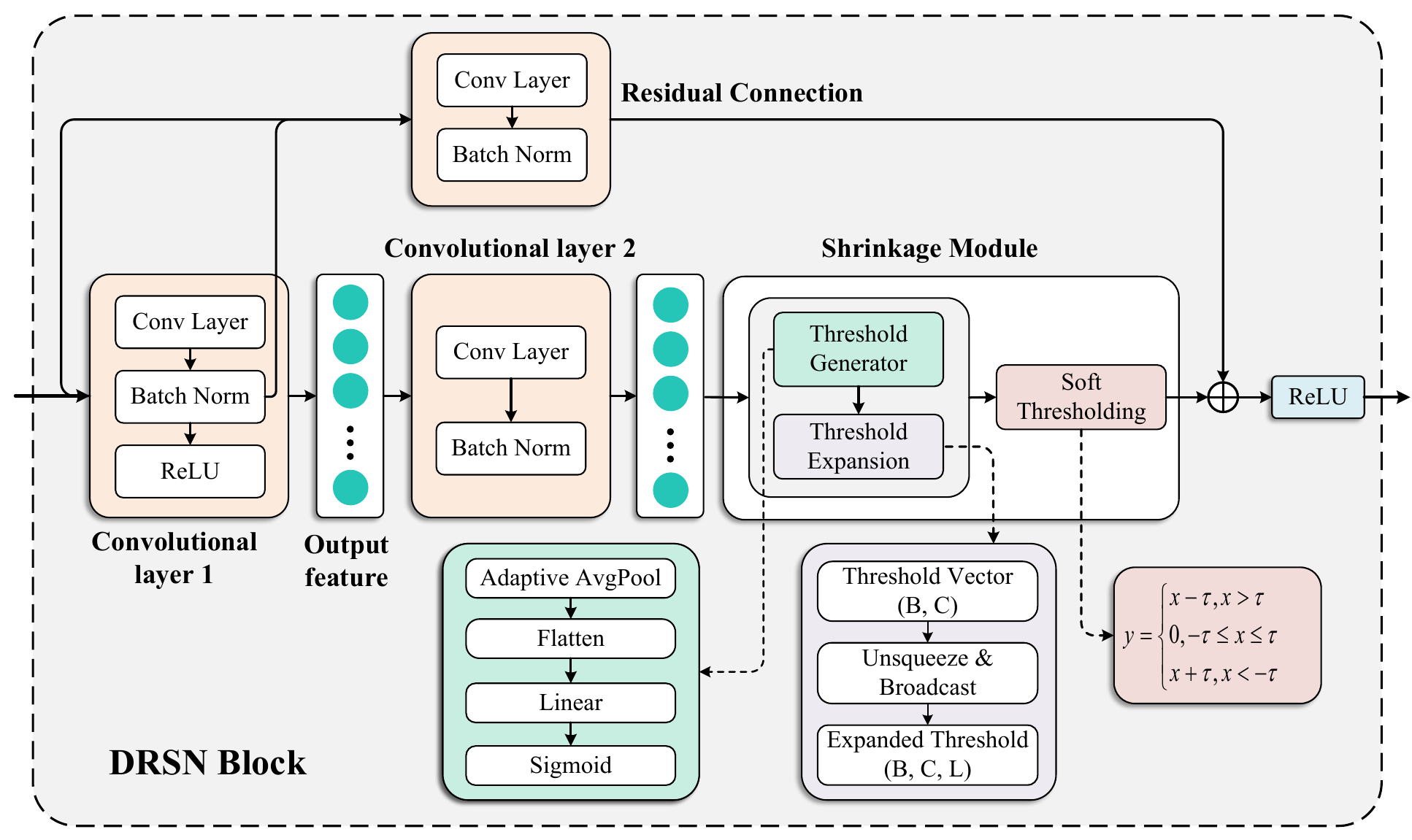}
\caption{Architecture of the DRSN module with convolution, soft-thresholding, and residual connections.}
\label{fig:drsn_architecture}
\end{figure}

Combining the core architectural concepts of CNNs and residual networks (ResNets), the DRSN effectively combines local feature extraction, residual learning, and adaptive denoising.  As illustrated in Fig.~\ref{fig:drsn_architecture}, the DRSN block consists of two convolutional layers, a soft-thresholding based shrinkage module, and a residual connection. This design not only enhances the expressive power of deep networks but also alleviates the vanishing gradient problem and improves robustness under noisy conditions.

The convolutional layers capture local temporal dependencies crucial for battery degradation modeling by sliding learnable filters over the input sequence. Stacking two convolutional layers allows the network to learn hierarchical and increasingly abstract features. Each convolutional layer is followed by batch normalization (BN) to stabilize and accelerate training, and a rectified linear unit (ReLU) activation function to introduce non-linearity and enhance model capacity. The convolution operation can be formulated as:
\begin{equation}
Y = X \cdot W + b,
\end{equation}
where \(X\) is the input, \(W\) denotes convolutional weights, and \(b\) is the bias term.

To suppress inherent noise and irrelevant variations in sensor data, DRSN employs an adaptive soft-thresholding function defined as:
\begin{equation}
Y = \mathrm{sign}(X) \cdot \max(|X| - \lambda, 0) =
\begin{cases}
X - \lambda, & X > \lambda, \\
0, & |X| \leq \lambda, \\
X + \lambda, & X < -\lambda,
\end{cases}
\end{equation}
where \(\lambda\) is a learnable threshold parameter. This smooth shrinkage function selectively attenuates small-magnitude features that likely correspond to noise while preserving important signals. Compared to hard-thresholding, the soft-thresholding offers stable gradients and avoids discontinuities during training.

The threshold \(\lambda\) is dynamically generated by a dedicated threshold generator, enabling channel-wise adaptive denoising. The generator first applies global average pooling over the temporal dimension on the output \(F_2\) of the second convolutional layer:
\begin{equation}
Z = \mathrm{AvgPool1d}(F_2) \in \mathbb{R}^{B \times C},
\end{equation}
where \(B\) and \(C\) denote the batch size and number of channels, respectively. Then, \(Z\) is passed through a two-layer fully connected network with ReLU and Sigmoid activations to produce normalized thresholds:
\begin{equation}
\lambda = \sigma\big(W_2 \cdot \mathrm{ReLU}(W_1 Z + b_1) + b_2 \big),
\end{equation}
where \(W_1, W_2, b_1, b_2\) are learnable parameters, and \(\sigma(\cdot)\) denotes the Sigmoid function. This design captures complex nonlinear inter-channel dependencies and constrains the thresholds within the \([0,1]\) range, adapting denoising strength dynamically based on input features.

Since \(\lambda \in \mathbb{R}^{B \times C}\) represents per-channel scalars, it is expanded along the temporal dimension to match the feature map size:
\begin{equation}
\lambda_{\mathrm{expanded}} = \lambda \otimes \mathbf{1}_L \in \mathbb{R}^{B \times C \times L},
\end{equation}
where \(\mathbf{1}_L\) is an all-ones vector of length \(L\). The soft-thresholding is then applied element-wise:
\begin{equation}
\widetilde{F}_2 = \mathrm{sign}(F_2) \cdot \max\big(|F_2| - \lambda_{\mathrm{expanded}}, 0\big).
\end{equation}

To ensure stable gradient propagation and preserve original feature information, DRSN employs residual connections that add the input \(X\) to the transformed features \(F(X)\):
\begin{equation}
Y = F(X) + X,
\end{equation}
where \(F(X)\) denotes the composite operations including convolution, normalization, activation, and soft-thresholding. When input and output channels differ (\(C_{in} \neq C_{out}\)), a \(1 \times 1\) convolution with batch normalization is applied to align dimensions:
\begin{equation}
\mathrm{Shortcut}(X) =
\begin{cases}
X, & C_{in} = C_{out}, \\
\mathrm{BN}_s(\mathrm{Conv}_{1 \times 1}(X)), & \text{otherwise}.
\end{cases}
\end{equation}
Finally, the output is obtained by applying a ReLU activation to the sum of the denoised features and the shortcut connection:
\begin{equation}
Y = \mathrm{ReLU}\big(\widetilde{F}_2 + \mathrm{Shortcut}(X)\big).
\end{equation}

By integrating convolutional feature extraction, adaptive soft-thresholding, and residual connections, the DRSN module achieves a powerful synergy that enhances feature representation while effectively suppressing noise and redundancy. The convolutional layers capture rich local patterns, the soft-thresholding adaptively removes irrelevant and noisy components in a data-driven manner, and the residual connections ensure stable gradient flow and efficient training even in deep architectures.Benefiting from this design, DRSN demonstrates robust and stable performance when handling noisy or partially missing sensor data in various real-world scenarios. Its feature shrinkage and residual connection mechanisms enhance generalization ability, providing a reliable foundation for practical applications such as battery health estimation and remaining useful life prediction.

\subsubsection{Transformer}
\begin{figure}[ht]
  \centering
  \begin{subfigure}{0.5\linewidth}
    \centering
    \includegraphics[width=\linewidth]{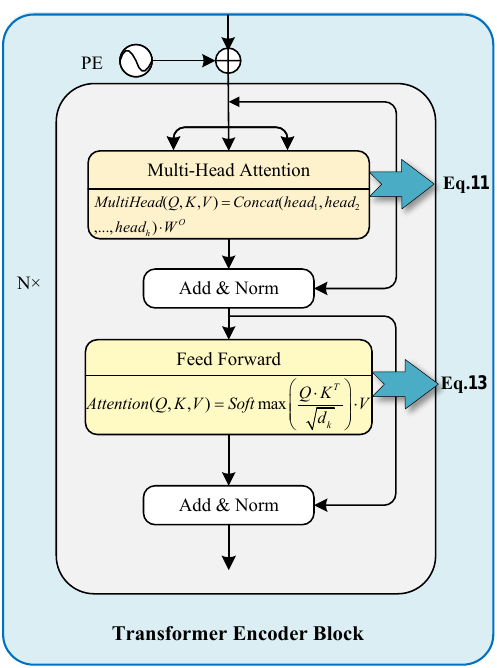}
    \caption{Transformer encoder block}
    \label{fig:transformer_arch}
  \end{subfigure}%
  \hfill  
  \begin{subfigure}{0.5\linewidth}
    \centering
    \includegraphics[width=\linewidth]{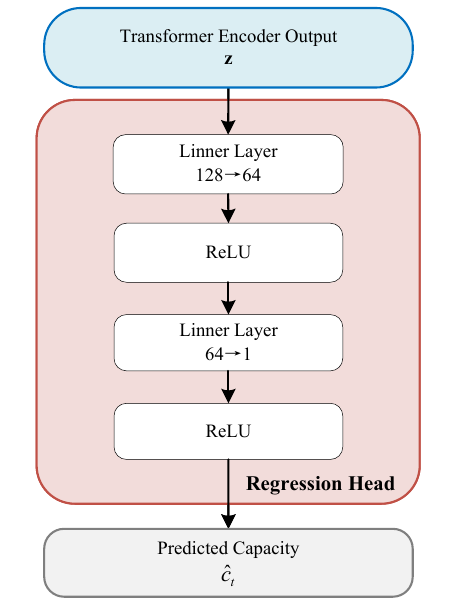}
    \caption{Regression head architecture}
    \label{fig:regression_head}
  \end{subfigure}

  \caption{Two architectures: (a) Transformer encoder block with multi-head attention and feed-forward layers; (b) Regression head.}
  \label{fig:two_arch}
\end{figure}

The Transformer architecture employs self-attention mechanisms to effectively capture long-range dependencies in time-series data, which is particularly advantageous for processing high-dimensional and lengthy sequences. A standard Transformer block consists of multiple stacked encoder layers, each integrating a multi-head self-attention module, residual connections, and a position-wise feedforward network, forming a robust and scalable architecture. The input sequence is first processed by the multi-head attention mechanism, which models dependencies between different positions within the sequence. Residual connections and layer normalization are employed to stabilize the training dynamics, followed by the feedforward network performing non-linear feature transformations to enhance the model's expressive power. As illustrated in Fig.~\ref{fig:two_arch}(a).

This architecture allows the Transformer to effectively learn global dependencies in time-series data, making it well-suited for analyzing complex sequential patterns. The self-attention mechanism enables the model to selectively focus on the most relevant parts of the sequence, thereby enhancing its predictive accuracy.

Multi-head self-attention extends the representational capacity of the model by computing attention in parallel across multiple heads. The formulation is defined as follows:
\begin{equation}
\text{MultiHead}(Q, K, V) = \text{Concat}(\text{head}_1, \text{head}_2, \dots, \text{head}_h) \cdot W^O.
\label{eq:multihead}
\end{equation}

Each attention head is computed individually, as follows:
\begin{equation}
\text{head}_i = \text{Attention}(Q \cdot W_i^Q, K \cdot W_i^K, V \cdot W_i^V). 
\label{eq:head}
\end{equation}

The scaled dot-product attention mechanism is defined as:
\begin{equation}
\text{Attention}(Q, K, V) = \text{Softmax}\left(\frac{Q \cdot K^T}{\sqrt{d_k}}\right) \cdot V. 
\label{eq:attention}
\end{equation}

Following the self-attention module, a position-wise feedforward network is applied to further enhance the non-linear transformation capacity, as follows:
\begin{equation}
Y = \text{ReLU}(X \cdot W_1 + b_1) \cdot W_2 + b_2. 
\label{eq:ffn}
\end{equation}

Here, \(Q\), \(K\), and \(V\) represent the query, key, and value matrices, respectively. \(W_i^Q\), \(W_i^K\), and \(W_i^V\) are the learned projection matrices for the \(i\)-th attention head, and \(W^O\) is the output projection matrix. The term \(d_k\) denotes the dimensionality of the key vectors. \(X\) is the output from the multi-head attention layer, and \(W_1\), \(W_2\), \(b_1\), \(b_2\) are the weights and biases of the feedforward layer.

\subsubsection{Regression Head}
The final stage of CDFormer is the Regression head, which maps the learned feature representations to a scalar output representing the predicted battery capacity at a future time step. This module enables the model to translate complex spatiotemporal features into a continuous estimate of capacity degradation. The architecture of the regression head is illustrated in Fig.~\ref{fig:two_arch}(b).

Let $\mathbf{z} \in \mathbb{R}^{d}$ denote the final hidden representation output by the Transformer encoder, typically corresponding to the last time step, i.e., $\mathbf{z} = \mathbf{h}_{T}$. The regression head is implemented using two linear layers with a ReLU activation in between:

\begin{equation}
\hat{c}_t = \mathbf{W}_2 \cdot \mathrm{ReLU}(\mathbf{W}_1 \cdot \mathbf{z} + \mathbf{b}_1) + \mathbf{b}_2,
\label{eq:regression_head}
\end{equation}
where $\mathbf{W}_1 \in \mathbb{R}^{d' \times d}$ and $\mathbf{b}_1 \in \mathbb{R}^{d'}$ represent the weight matrix and bias vector of the first linear layer, respectively. Similarly, $\mathbf{W}_2 \in \mathbb{R}^{1 \times d'}$ and $\mathbf{b}_2 \in \mathbb{R}$ correspond to the parameters of the second linear layer. The output $\hat{c}_t \in \mathbb{R}$ denotes the predicted capacity at a future time step $t$.

The learned feature vector $\mathbf{z}$ is first transformed by a linear layer followed by a ReLU activation, and then passed through a second linear layer to produce the scalar capacity prediction $\hat{c}_t$. To avoid physically implausible negative predictions, a ReLU activation is applied to the regression output in the smaller, noisier NASA dataset to stabilize training. In contrast, the larger and cleaner CALCE dataset omits this constraint, allowing greater regression flexibility.

This architecture allows the model to learn a nonlinear mapping from high-level temporal features to future capacity values. The predicted capacity sequence can be used to determine the EOL point by identifying when it drops below $70\%$ of the rated capacity.

\subsection{Temporal Data Augmentation Techniques}
To improve the model's generalization capability and robustness against diverse degradation patterns, especially under the complex conditions typical of real-world operating environments, we propose a composite temporal data augmentation strategy for the training phase. This approach incorporates three techniques: time warping, time resampling, and Gaussian noise perturbation, among which the former two are custom-designed methods developed in this work. Details of temporal data augmentation are shown in Fig.~\ref{fig:temporal data augmentation}:
\begin{figure}[htbp]
    \centering
    \includegraphics[width=1\linewidth]{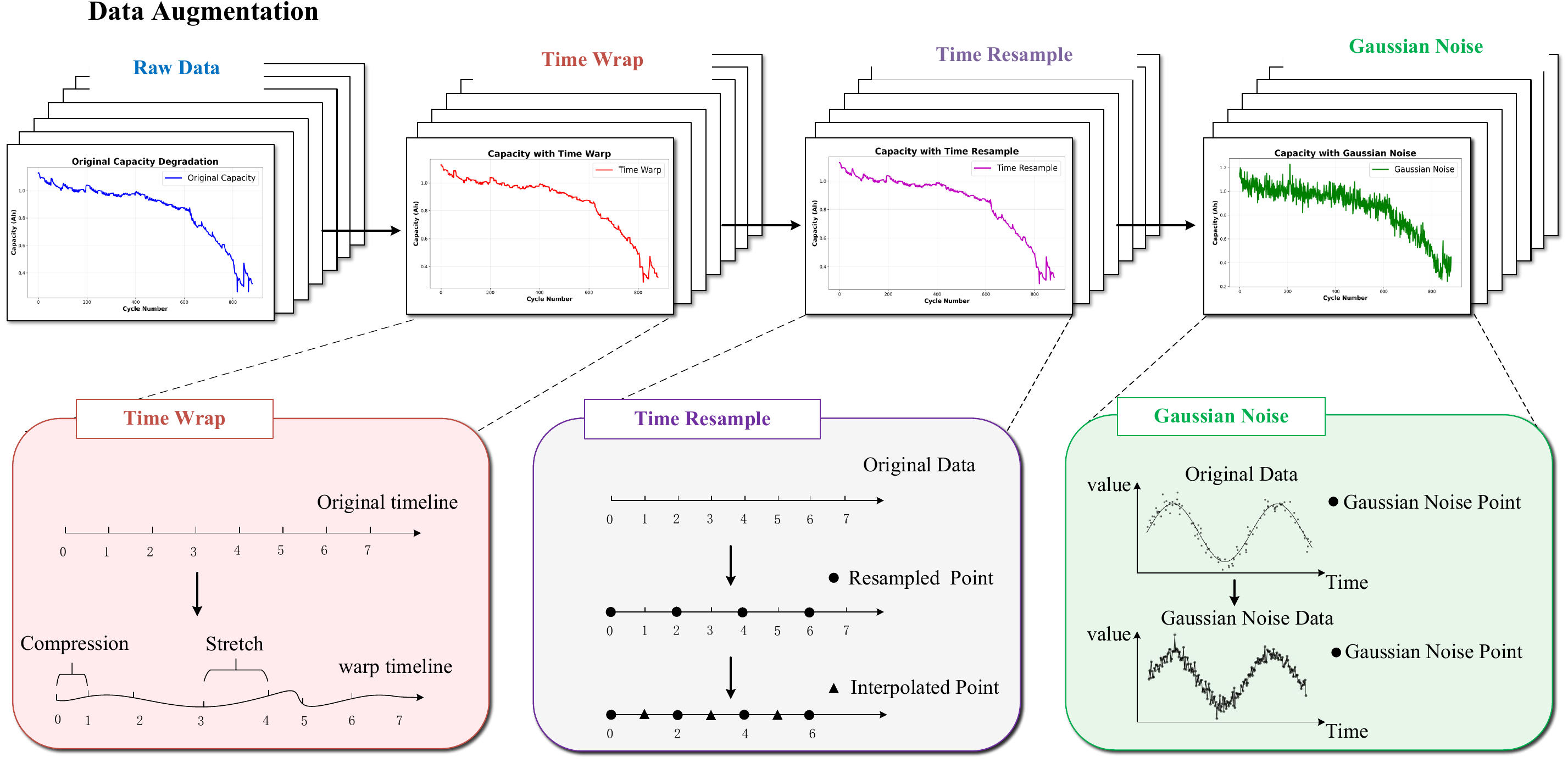}
    \caption{Three Temporal Augmentation Techniques}
    \label{fig:temporal data augmentation}
\end{figure}

\subsubsection{Time Warping}
To simulate temporal distortions commonly observed in real-world scenarios, one effective augmentation method is time warping, which applies nonlinear transformations to the time axis of a sequence. By introducing small perturbations along the time axis, this method enhances the model's robustness to temporal variations and improves its generalization performance in time-series tasks.
Formally, we define the original time series as follows:
\begin{equation}
\chi_0 = \{x_1, x_2, \dots, x_T\}, \quad (t = i \Delta t).
\label{eq:original_time_series}
\end{equation}
Accordingly, we construct a new set of time indices as follows:
\begin{equation}
D = \{d_1, d_2, \dots, d_T\}.
\label{eq:time_indices}
\end{equation}
Here, each time index \(d_t\) is derived by perturbing its original position \(t\), as follows:
\begin{equation}
d_t = t_i + \delta_i, \quad \delta_i \sim U(-\alpha \Delta t, \alpha \Delta t),
\label{eq:time_perturbation}
\end{equation}
where the \(\alpha\) is a perturbation strength hyperparameter, and \(\Delta t\) is the time resolution.
A new time series \(\chi'\) is generated by linearly interpolating the original sequence \(\chi_0\) at the perturbed time points \(D\), as follows:
\begin{equation}
\chi' = \text{Interp}(\chi_0, D).
\label{eq:interpolated_series}
\end{equation}

\subsubsection{Time Resampling}
Time resampling augments temporal data by randomly selecting a subset of time points from the original time series and interpolating them to restore the original sequence length, thereby generating a new time-series sample. This approach not only increases data diversity but also enhances the model’s ability to learn critical temporal patterns, improving its sensitivity to key features and boosting overall predictive performance.

In addition, a time resampling method can be used to further augment the data. This method randomly selects a portion of the time points from the original sequence and interpolates them back to the original length. Specifically, \(n = \lfloor \rho T \rfloor\) time points (where \(\rho \in [0,1]\) is the sampling ratio) are randomly selected from \(\chi_0\), sorted in ascending order, and then interpolated to generate a new time series \(\chi'\), as follows:
\begin{equation}
\chi' = \text{Interp}(\text{Sample}(\chi_0, \rho T)).
\label{eq:time_resampling}
\end{equation}

\subsubsection{Gaussian Noise}
In addition to the temporal augmentation methods described above, Gaussian noise perturbation is employed as a simple yet effective augmentation technique. This method adds random Gaussian-distributed noise to the original time series, simulating measurement fluctuations and sensor noise commonly encountered in practical data acquisition.

Formally, given the original time series \(\chi_0 = \{x_1, x_2, \dots, x_T\}\), the augmented sequence \(\chi''\) is obtained by adding Gaussian noise \(\epsilon_t\) sampled from \(\mathcal{N}(0, \sigma^2)\) independently to each time point:

\begin{equation}
\chi'' = \{x_1 + \epsilon_1, x_2 + \epsilon_2, \dots, x_T + \epsilon_T\}, \quad \epsilon_t \sim \mathcal{N}(0, \sigma^2),
\label{eq:gaussian_noise}
\end{equation}
where the \(\sigma\) is a hyperparameter controlling the noise intensity. The addition of Gaussian noise effectively enhances the diversity of the training data by simulating realistic signal perturbations, thereby improving the model’s robustness and generalization ability.

The three temporal data augmentation methods collectively increase the diversity of training data, improving the model’s robustness and generalization to complex conditions. This is particularly important for lithium-ion batteries in real-world applications, where irregular usage patterns, measurement noise, and external disturbances are common.
By simulating such variations, these augmentations help the model deliver more stable and accurate capacity predictions. Ablation studies later quantify their individual and combined effects, confirming their contribution to enhanced performance.

\section{Case Study}
\subsection{Datasets}
This study employs two publicly available lithium-ion battery degradation datasets: the NASA Prognostics Center of Excellence (NASA PCoE) dataset and the Center for Advanced Life Cycle Engineering (CALCE) dataset. The NASA dataset comprises four batteries tested under controlled laboratory conditions with full charge–discharge cycles, while the CALCE dataset includes four batteries tested at a constant ambient temperature but with varying charging and discharging protocols. Detailed information on battery types, nominal capacities, sampling frequencies, and testing conditions is provided in Tables~\ref{tab:nasa_battery_detailed} and \ref{tab:calce_battery_detailed}. These two datasets are widely adopted benchmarks in the field of RUL prediction, serving as standard references for performance evaluation. The NASA dataset emphasizes capacity degradation patterns under stable conditions, whereas the CALCE dataset introduces more representative operational variability through diverse charging–discharging strategies, collectively covering a wide range of degradation modes and usage scenarios. 
\begin{table}[htbp]
\centering
\caption{\\Detailed Information of NASA Lithium-Ion Battery Dataset}
\label{tab:nasa_battery_detailed}
\begin{tabularx}{\linewidth}{l X}
\toprule
\textbf{Parameter} & \textbf{Description / Value} \\
\midrule
Battery Type & 18650 Cylindrical Li-ion Cell \\
Manufacturer & Panasonic  \\
Rated Capacity & 2.0 Ah \\
Nominal Voltage & 3.6 V \\
Charge Mode & CC-CV 1.5 A constant current cutoff voltage 4.2 V \\
Discharge Cutoff Voltage & B0005: 2.7 V; B0006: 2.5 V; B0007: 2.2 V; B0018: 2.5 V \\
Discharge Current & 2 A \\
Charge Cutoff Condition & Constant voltage at 4.2 V current taper to 0.02 A \\
Cycle Count & B0005: 168; B0006: 168; B0007: 168; B0018 132 \\
Sampling Frequency & Irregular, approximately 1 Hz during discharge (varies across cycles) \\
Measured Signals & Voltage V Current A Temperature ℃ Time s Internal Resistance m$\Omega$ Cycle Index \\
Operating Temperature & 24 $\pm$ 1 ℃ \\
Environmental Conditions & Laboratory controlled temperature \\
Cycle Profile & Standard charge discharge cycles \\
Data Format & MATLAB .mat files with time series sensor data per cycle \\
\bottomrule
\end{tabularx}
\end{table}

\begin{table}[htbp]
\centering
\caption{\\Detailed Information of CALCE Lithium-Ion Battery Dataset}
\label{tab:calce_battery_detailed}
\begin{tabularx}{\linewidth}{l X}
\toprule
\textbf{Parameter} & \textbf{Description / Value} \\
\midrule
Battery Type & Lithium-ion pouch cell \\
Manufacturer & Not explicitly specified varies \\
Rated Capacity & 1.1 Ah nominal \\
Nominal Voltage & 3.7 V \\
Charge Mode & CC-CV 0.5C constant current cutoff voltage 4.2 V \\
Discharge Cutoff Voltage & 2.7 V \\
Discharge Current & 1C 1.1 A \\
Charge Cutoff Condition & Constant voltage charge at 4.2 V until current tapers to 0.05A \\
Cycle Count & CS2\_35: approximately 900; CS2\_36: approximately 950; CS2\_37: approximately 1000; CS2\_38: approximately 1000 \\
Sampling Frequency & 1 Hz \\
Measured Signals & Voltage V Current A Temperature ℃ Charge Capacity Ah Discharge Capacity Ah Internal Resistance m$\Omega$ Time s Cycle Index \\
Operating Temperature & 25 $\pm$ 2 ℃ \\
Environmental Conditions & Laboratory controlled temperature \\
Cycle Profile & Standard CC-CV charge and constant 1C discharge cycles \\
Data Format & Multiple time ordered Excel files recording detailed cycling data \\
\bottomrule
\end{tabularx}
\end{table}

\subsection{Implement Details}
During the training process, a LOOCV scheme was adopted, where data from one battery served as the test set in each iteration while the remaining batteries’ data were utilized for training. This strategy ensures comprehensive evaluation across the full battery life cycle, thereby improving the model’s generalization capability and predictive reliability.

Considering the differences in scale and feature distribution between the NASA and CALCE datasets, input features and hyperparameters were tailored accordingly. For the relatively smaller NASA dataset, the input features include average voltage (Voltage\_avg), average current (Current\_avg), average temperature (Temp\_avg), and discharge capacity (Capacity). In contrast, the larger and more complex CALCE dataset uses discharge capacity (Capacity), constant current charge time, and state of health (SOH) as inputs. The prediction target for both datasets remains the battery discharge capacity.

Model training employed the Huber loss function optimized via the Adam algorithm. The learning rates were set at 0.0005 for the NASA dataset and 0.001 for the CALCE dataset, with exponential decay rates $\beta_1 = 0.9$ and $\beta_2 = 0.999$ for the first and second moment estimates, respectively. To mitigate overfitting, early stopping with a patience of 15 epochs and L2 regularization (weight decay coefficient 0.001) were applied. Batch size was fixed at 32, with maximum epochs set to 200 for NASA and 500 for CALCE.

As summarized in Table~\ref{tab:input_features}, different sets of input features were used for the NASA and CALCE datasets based on their characteristics. Features with \checkmark\ indicate direct usage in model training, while features with descriptions in parentheses are derived through post-processing or calculation.
\begin{table}[htbp]
\centering
\caption{\\Input features used in the NASA and CALCE datasets.}
\label{tab:input_features}
\begin{tabularx}{\linewidth}{>{\raggedright\arraybackslash}l
                             >{\centering\arraybackslash}X
                             >{\raggedleft\arraybackslash}l}
\toprule
\textbf{Input Feature} & \textbf{NASA} & \textbf{CALCE} \\
\midrule
\makecell[l]{Average voltage  (Voltage\_avg)} & \checkmark & \\
\makecell[l]{Average current  (Current\_avg)} & \checkmark & \\
\makecell[l]{Average temperature (Temp\_avg)} & \checkmark & \\
\makecell[l]{Discharge capacity  (Capacity)} & \checkmark & \checkmark \\
\makecell[l]{Constant-current charge time  (derived from CC-phase duration)} & & \checkmark \\
\makecell[l]{State of health (SOH)  \(\left(SOH = \frac{C_{\text{current}}}{C_{\text{initial}}} \times 100\%\right)\)} & & \checkmark \\
\bottomrule
\end{tabularx}
\end{table}

\subsection{Metrics}
To comprehensively evaluate the performance of the proposed method, we follow prior studies and employ several commonly used evaluation metrics, including root mean square error (RMSE), mean absolute error (MAE), and relative error (RE). These metrics are standard quantitative indicators in deep learning-based life prediction tasks. They assess the discrepancy between predicted and true values from different perspectives, while also reflecting prediction performance under practical sensing conditions in terms of accuracy, stability, and reliability.

\noindent\textbf{RMSE} measures the square root of the average squared difference between the predicted and actual values. It is defined as:
\begin{equation}
RMSE = \sqrt{\frac{1}{n} \sum_{i=1}^{n} (y_{\text{true},i} - y_{\text{pred},i})^2}.
\label{eq:RMSE}
\end{equation}
Here \(y_{\text{true},i}\) is the ground truth capacity of the \(i\)-th sample, \(y_{\text{pred},i}\) is the corresponding predicted value, and \(n\) is the total number of samples.

\noindent\textbf{MAE} quantifies the average absolute deviation between predicted values and true values, providing an intuitive measure of prediction error. It is computed as follows:
\begin{equation}
MAE = \frac{1}{n} \sum_{i=1}^{n} |y_{\text{true},i} - y_{\text{pred},i}|. 
\label{eq:MAE}
\end{equation}
Here, \(n\) denotes the number of samples, and \(y_{\text{true},i}\), \(y_{\text{pred},i}\) represent the true and predicted values of the \(i\)-th sample, respectively.

\noindent\textbf{RE} quantifies the relative deviation between the predicted and actual values, making it suitable for evaluating prediction accuracy across data with varying scales. In this study, RE is calculated based on the cycle number at the EOL point, as follows:
\begin{equation}
RE = \frac{|N_{\text{true}}^{\text{EOL}} - N_{\text{pred}}^{\text{EOL}}|}{N_{\text{true}}^{\text{EOL}}}. 
\label{eq:RE}
\end{equation}
Here, \( N_{\text{true}}^{\text{EOL}} \) and \( N_{\text{pred}}^{\text{EOL}} \) refer to the cycle numbers at which the true and predicted capacities first fall below 70\% of the rated capacity \(C\), respectively.

In this paper, RMSE, MAE, and RE are adopted as the primary quantitative metrics to evaluate the performance of various models. Beyond these quantitative assessments, we also perform qualitative analyses in the experimental section, such as visualizations and trend comparisons. These complementary evaluations provide further evidence of the effectiveness and reliability of the proposed method when applied to real-world battery sensing data.

\section{Result and Discussion}
In this section, we present a comprehensive evaluation of the proposed CDFormer model. We first compare its performance against state-of-the-art methods and other deep learning-based variants we designed, followed by ablation studies to analyze the temporal data augmentation techniques.

\subsection{Comparison with state-of-the-art (SOTA) Methods}
\subsubsection{Comparison with Existing Methods}
To comprehensively evaluate the performance of CDFormer, we conduct a comparative study against a range of established models for RUL prediction of lithium-ion batteries. These include classical sequence modeling networks such as RNN, LSTM, and GRU, as well as more recent state-of-the-art architectures like DeTransformer, AttMoE, and SA-LSTM. It is noteworthy that all selected baseline models utilize discharge capacity as the sole input feature, reflecting the reliance on univariate inputs in some prior studies. In contrast, CDFormer is capable of flexibly integrating multiple input features, thereby more effectively mining the latent temporal information inherent in the battery degradation process and improving prediction performance.

\begin{table}[htbp]
\setlength{\tabcolsep}{3.5pt}
\centering
\caption{\\Quantitative evaluations of CDFormer and baseline models on the NASA and CALCE datasets.}
\label{tab:Comparison table with existing methods}
\begin{tabularx}{\linewidth}{l X X X X X X}
\toprule
\multirow{2}{*}{Models} & \multicolumn{3}{c}{\textit{NASA}} & \multicolumn{3}{c}{\textit{CALCE}} \\
\cmidrule(lr){2-4} \cmidrule(lr){5-7}
 & RMSE $\downarrow$ & MAE $\downarrow$ & RE $\downarrow$ & RMSE $\downarrow$ & MAE $\downarrow$ & RE $\downarrow$ \\
\midrule
RNN~\cite{Catelani2021} & 0.1435 & 0.1205 & 0.3692 & 0.0534 & 0.0423 & 0.1418 \\
LSTM~\cite{ZhaoLSTM2022} & 0.1573 & 0.1356 & 0.3692 & 0.0720 & 0.0487 & 0.1466 \\
GRU~\cite{Lin2023}  & 0.0863 & 0.0675 & 0.3002 & 0.0973 & 0.0756 & 0.1646 \\
DeTransformer~\cite{Chen2022} & 0.0863 & 0.0746 & 0.2609 & 0.0616 & 0.0481 & 0.1282 \\
SA-LSTM~\cite{WangLSTM2023} & 0.0849 & 0.0751 & 0.3725 & 0.0324 & 0.0263 & 0.1097 \\
AttMoE~\cite{ChenAttMoE2024} & 0.0847 & 0.0730 & 0.2849 & 0.0250 & 0.0180 & 0.1345 \\
CDFormer & \textbf{0.0671} & \textbf{0.0586} & \textbf{0.2365} & \textbf{0.0179} & \textbf{0.0106} & \textbf{0.0877} \\
\bottomrule
\end{tabularx}
\end{table}
Table~\ref{tab:Comparison table with existing methods} reports the prediction results on the NASA and CALCE datasets. The symbol $\downarrow$ indicates that lower values correspond to better performance. Bold numbers denote the best performance. CDFormer consistently achieved superior performance across all evaluation metrics. On the NASA dataset, it reduced RMSE, MAE, and RE by 20.8\%, 19.7\%, and 17.0\%, respectively, compared to the best-performing baseline, AttMoE. Even when compared with DeTransformer, which also relied on Transformer-based modeling, CDFormer demonstrated improvements of 22.2\%, 21.4\%, and 9.4\% in RMSE, MAE, and RE, respectively. On the CALCE dataset, CDFormer further enhanced its performance. Compared to the best-performing model AttMoE, it achieved error reductions of 28.4\% and 41.1\% in RMSE and MAE, respectively, while the RE value decreased by 20.1\% compared to the best-performing SA-LSTM. Although most models generally performed better on the CALCE dataset due to its larger sample size and more diverse degradation scenarios, the GRU model exhibited a performance drop. This was likely attributable to its simplified gating mechanism, which limited its ability to model complex degradation dynamics and long-term dependencies, thereby reducing its generalization capability. These results indicate that CDFormer robustly captures degradation dynamics in real measurement data.

\begin{figure}[htbp]
  \centering
  \begin{minipage}[t]{0.49\linewidth}
    \centering
    \includegraphics[width=\linewidth]{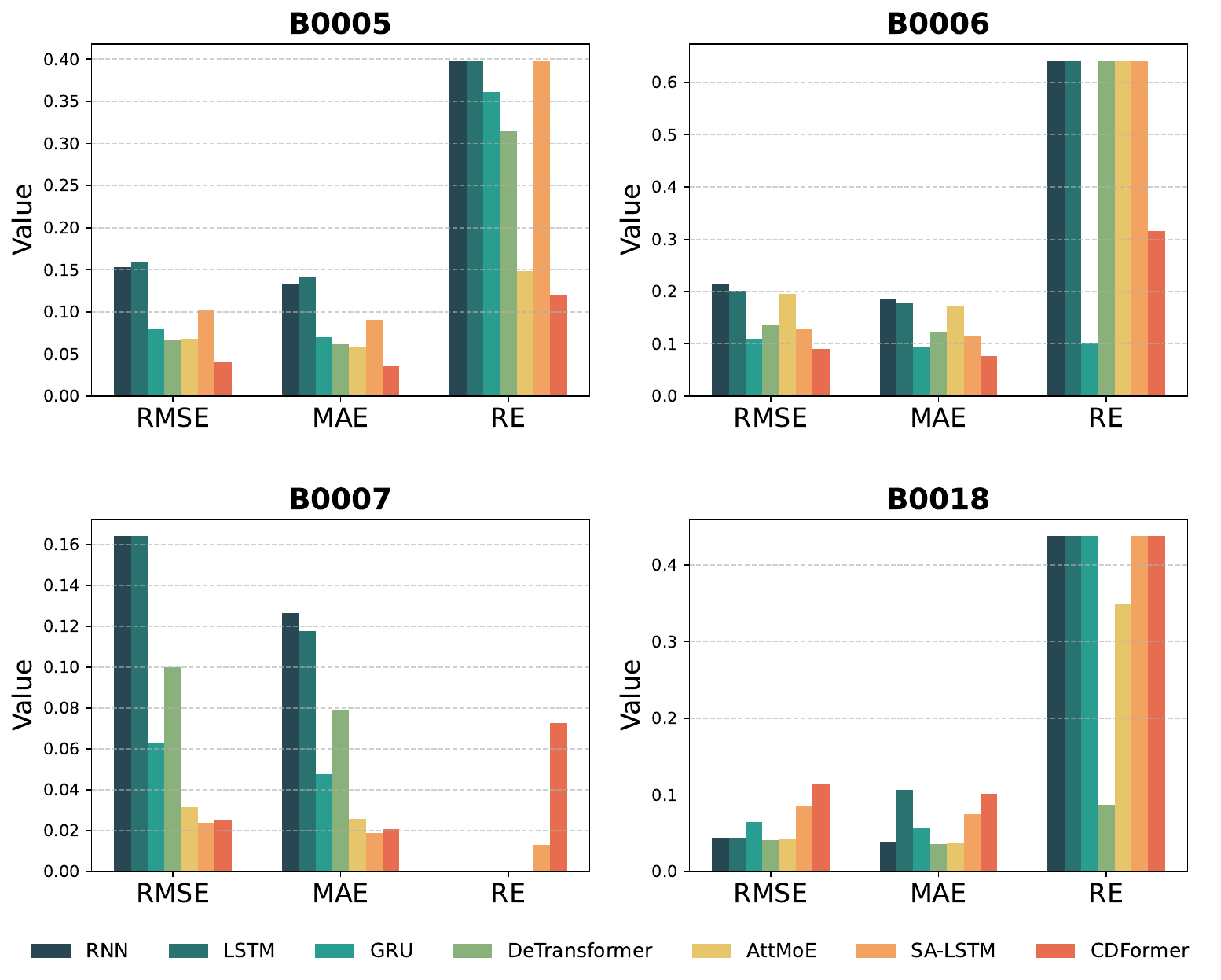}
    \includegraphics[width=\linewidth]{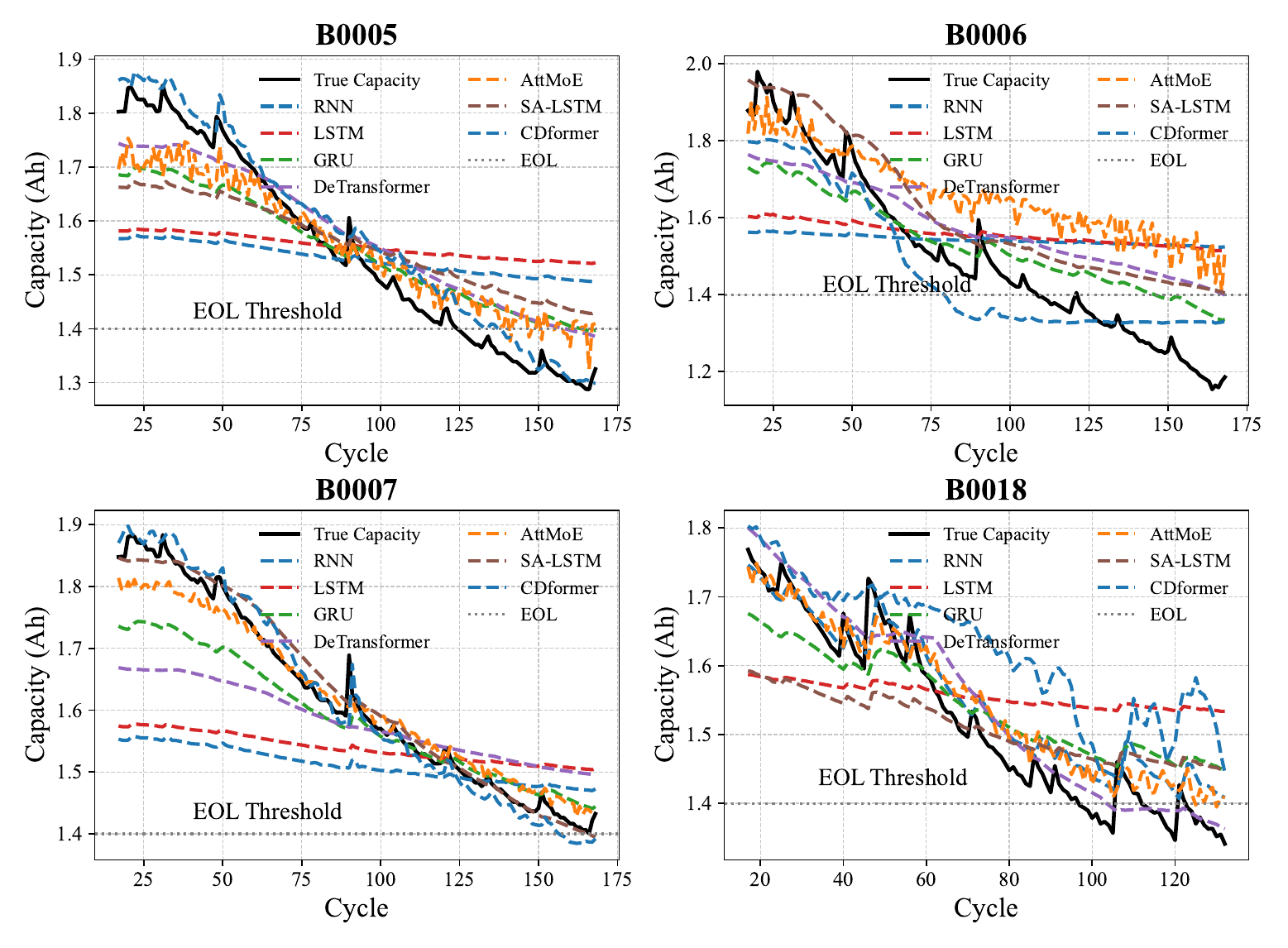}
    \caption*{(a) NASA dataset}
    \label{fig:combined_aligned_a}
  \end{minipage}
  \hspace{0.01\linewidth}
  \begin{minipage}[t]{0.49\linewidth}
    \centering
    \includegraphics[width=\linewidth]{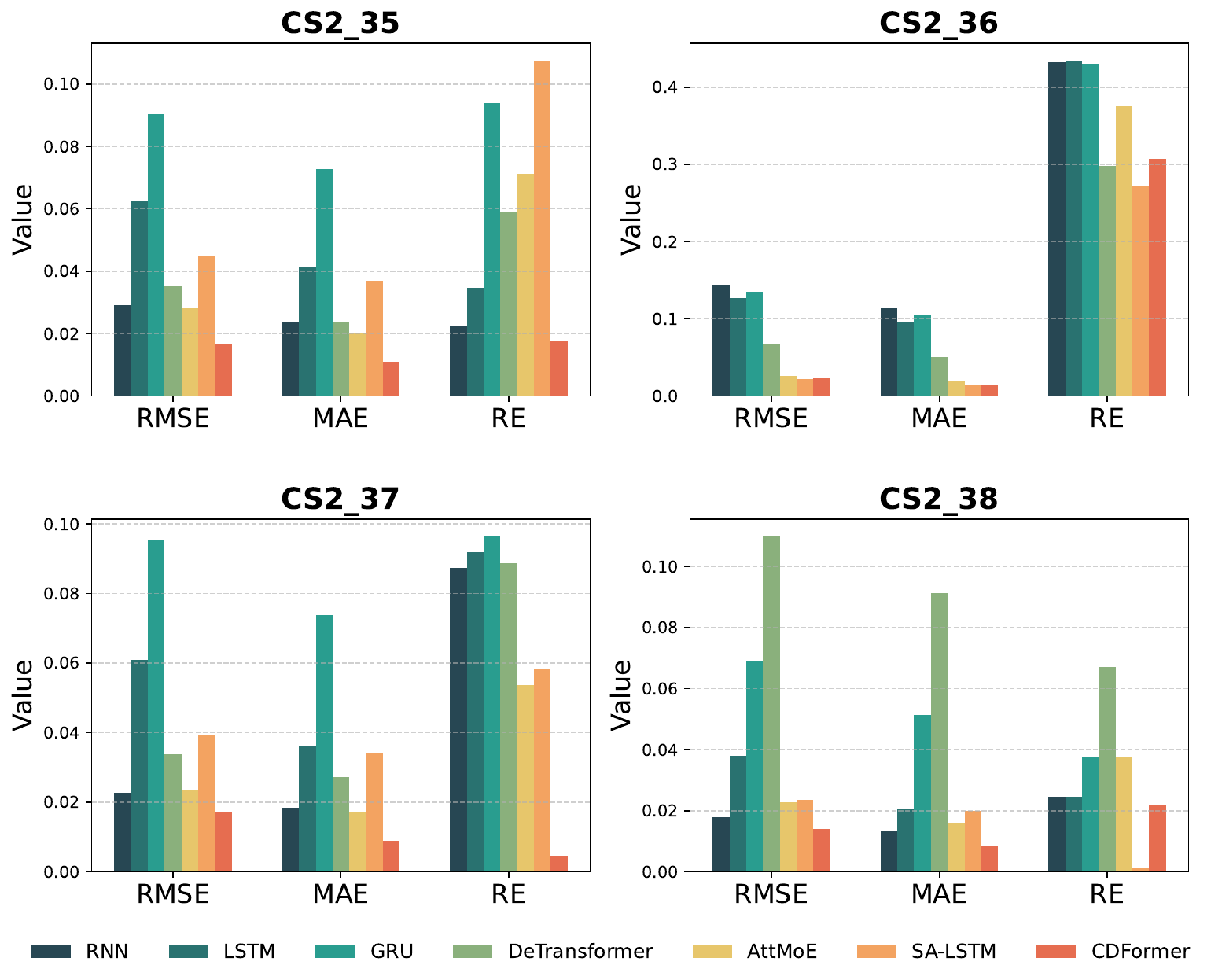}
    \includegraphics[width=\linewidth]{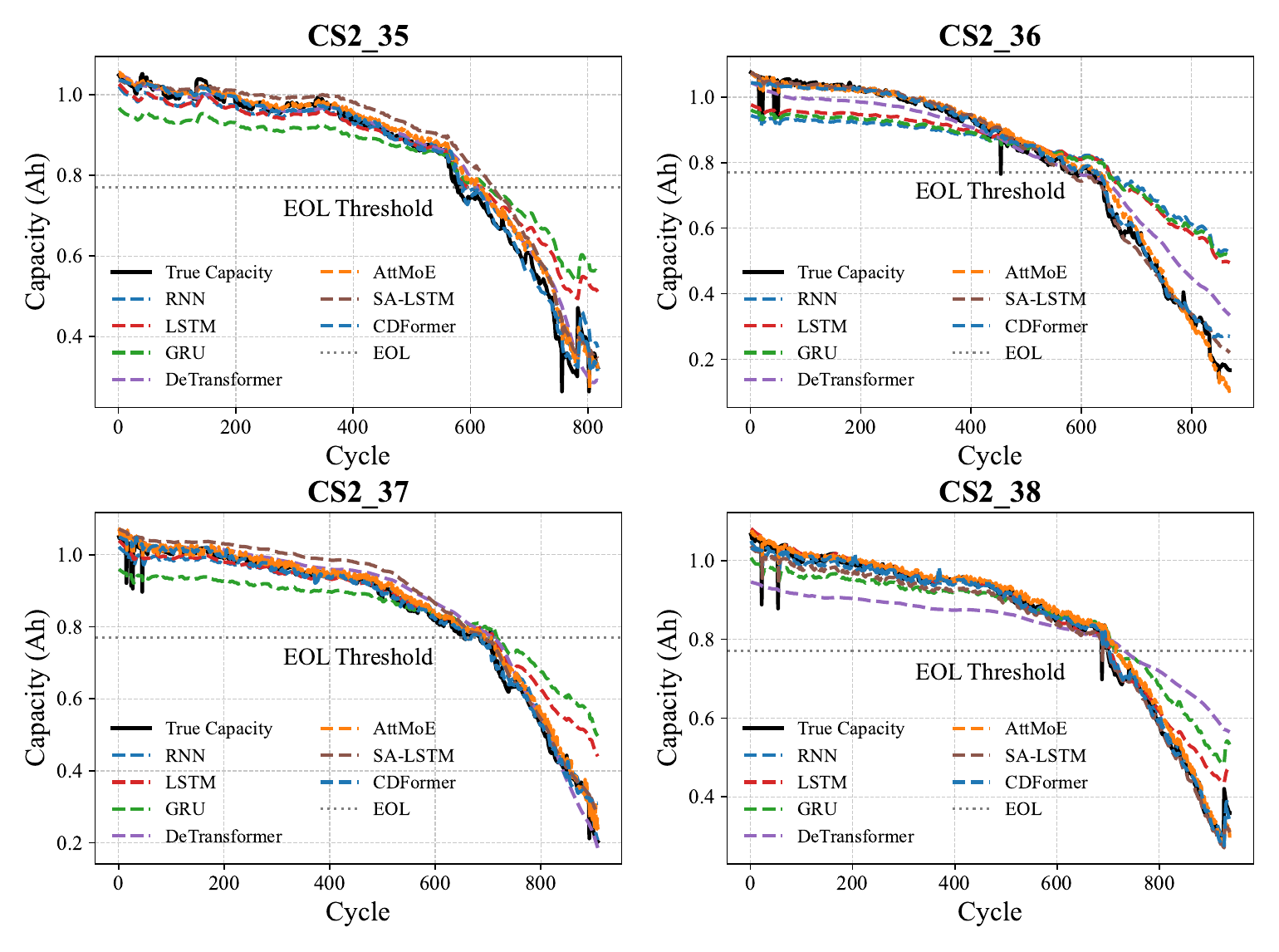}
    \caption*{(b) CALCE dataset}
    \label{fig:combined_aligned_b}
  \end{minipage}

  \caption{Evaluation of CDFormer and baseline models on NASA and CALCE datasets. 
  Left column: RMSE, MAE, and RE metrics (top) and predicted vs true capacity curves (bottom) for NASA. 
  Right column: same metrics and predicted vs true curves for CALCE.}
  \label{fig:combined_aligned}
\end{figure}
As shown in Fig.~\ref{fig:combined_aligned}, the left column (a) presents the results on the NASA dataset, and the right column (b) corresponds to the CALCE dataset. The upper subplots show RMSE, MAE, and RE metrics for each battery (B0005–B0018 and CS2\_35–CS2\_38). The lower subplots illustrate predicted versus true capacity degradation curves, highlighting how well the models track real degradation trends under measured conditions.

For the NASA dataset (Fig.~\ref{fig:combined_aligned}(a)), CDFormer achieved the lowest RMSE and MAE on B0005 and B0006. Although it did not perform best on B0007, the gap with the optimal SA-LSTM model was small. Although RE was not always the lowest, the differences were minor and offset by lower absolute errors. The degradation curves indicated that CDFormer closely followed the ground truth during stable phases and captured abrupt capacity changes, effectively tracking both rapid transitions and subtle local variations.

For the CALCE dataset (Fig.~\ref{fig:combined_aligned}(b)), CDFormer maintained stable performance across all batteries, frequently ranking first or near first. The predicted curves followed rapid capacity changes and local degradation patterns more closely than baseline models. The larger sample size and more diverse degradation scenarios in CALCE allowed the model to learn richer representations, supporting better generalization and more stable predictions.

\subsubsection{Comparison with Other Deep Learning-Based Variants}
To further verify the effectiveness of the proposed CDFormer architecture, we designed and evaluated three representative model variants with distinct structural configurations. The first variant, referred to as Baseline-FC, is a fully connected neural network composed of stacked dense layers, serving as a simple reference for modeling capacity degradation. The second variant, termed CNN-FC, integrates one-dimensional convolutional layers with fully connected layers to enhance the extraction of local temporal features. The third variant, named CNN-Trans, combines 1D convolutional layers with Transformer encoders, allowing the model to capture both short- and long-range temporal dependencies. These variants provide a comprehensive framework for analyzing the contributions of different architectural components—namely dense layers, CNNs, and Transformer blocks—to the overall prediction performance. CDFormer incorporates all three core modules, including 1D-CNN, DRSN, and Transformer encoders, and is thus expected to demonstrate enhanced learning capacity and generalization ability under complex degradation scenarios.

\begin{table}[htbp]
\setlength{\tabcolsep}{6pt}
\centering
\caption{\\Quantitative evaluations of CDFormer and other deep learning-based variant models.}
\label{tab:Comparison table with variant models}
\begin{tabularx}{\linewidth}{l X X X X X X}
\toprule
\multirow{2}{*}{Models} & \multicolumn{3}{c}{\textit{NASA}} & \multicolumn{3}{c}{\textit{CALCE}} \\
\cmidrule(lr){2-4} \cmidrule(lr){5-7}
& RMSE $\downarrow$ & MAE $\downarrow$ & RE $\downarrow$ & RMSE $\downarrow$ & MAE $\downarrow$ & RE $\downarrow$ \\
\midrule
Baseline-FC & 0.1952 & 0.1822 & 0.5572 & 0.0309 & 0.0184 & 0.0980 \\
CNN-FC & 0.1117 & 0.0981 & 0.2322 & 0.0209 & 0.0127 & \textbf{0.0828} \\
CNN-Transformer & 0.1059 & 0.0904 & \textbf{0.2264} & 0.0225 & 0.0133 & 0.0896 \\
CDFormer & \textbf{0.0671} & \textbf{0.0586} & 0.2365 & \textbf{0.0179} & \textbf{0.0106} & 0.0877 \\
\bottomrule
\end{tabularx}
\end{table}

Table~\ref{tab:Comparison table with variant models} reports the average RMSE, MAE, and RE metrics of each model over four batteries in the NASA and CALCE datasets, reflecting the overall prediction performance. CDFormer achieved the lowest RMSE and MAE values on both datasets, indicating superior absolute error performance. Although its RE metric was slightly higher than some competing models, the differences were minor and offset by significantly reduced absolute errors.

\begin{figure}[ht]
  \centering
  \begin{subfigure}{0.49\linewidth}
    \centering
    \includegraphics[width=\linewidth]{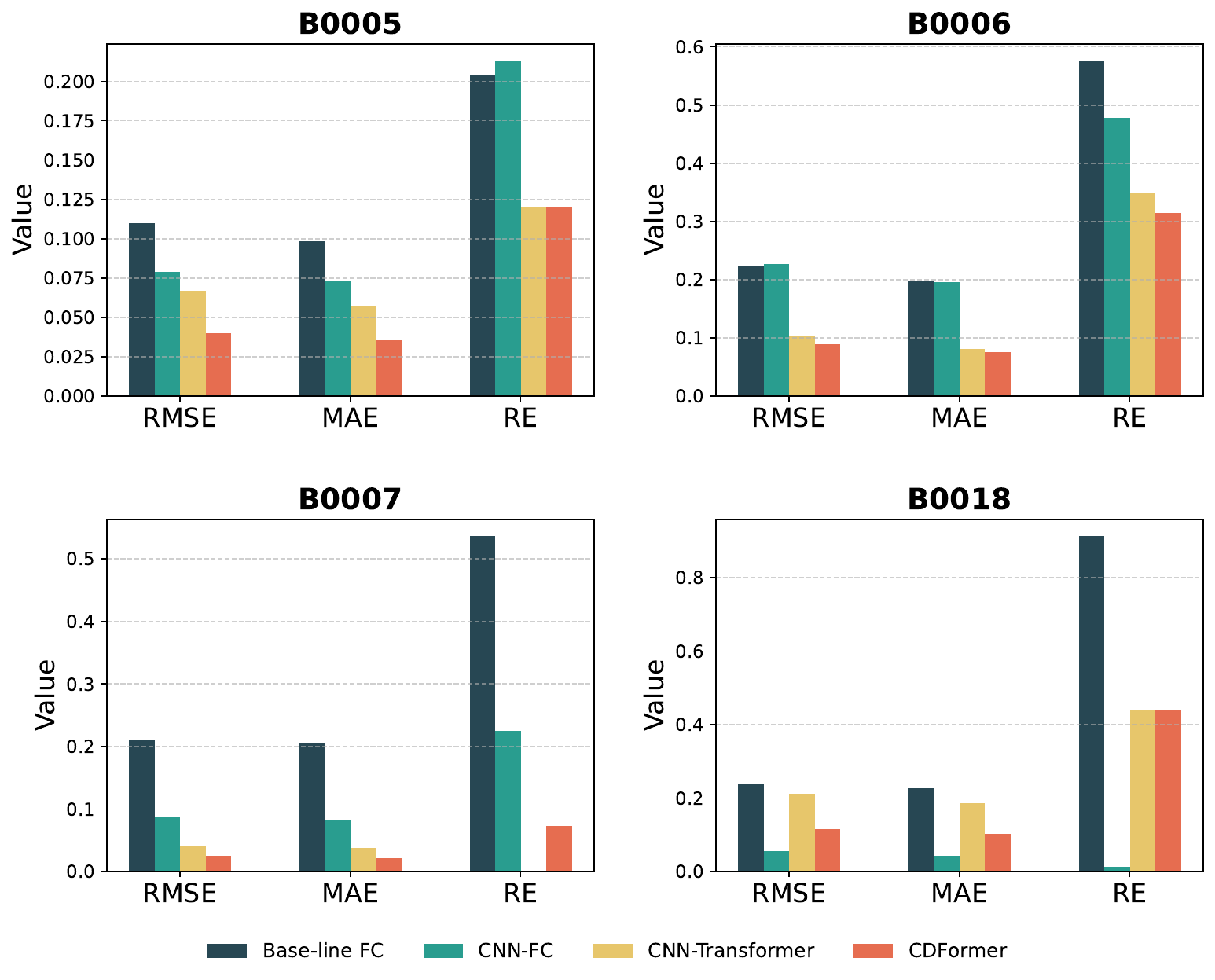} \\ % top: RMSE, MAE, RE
    \includegraphics[width=\linewidth]{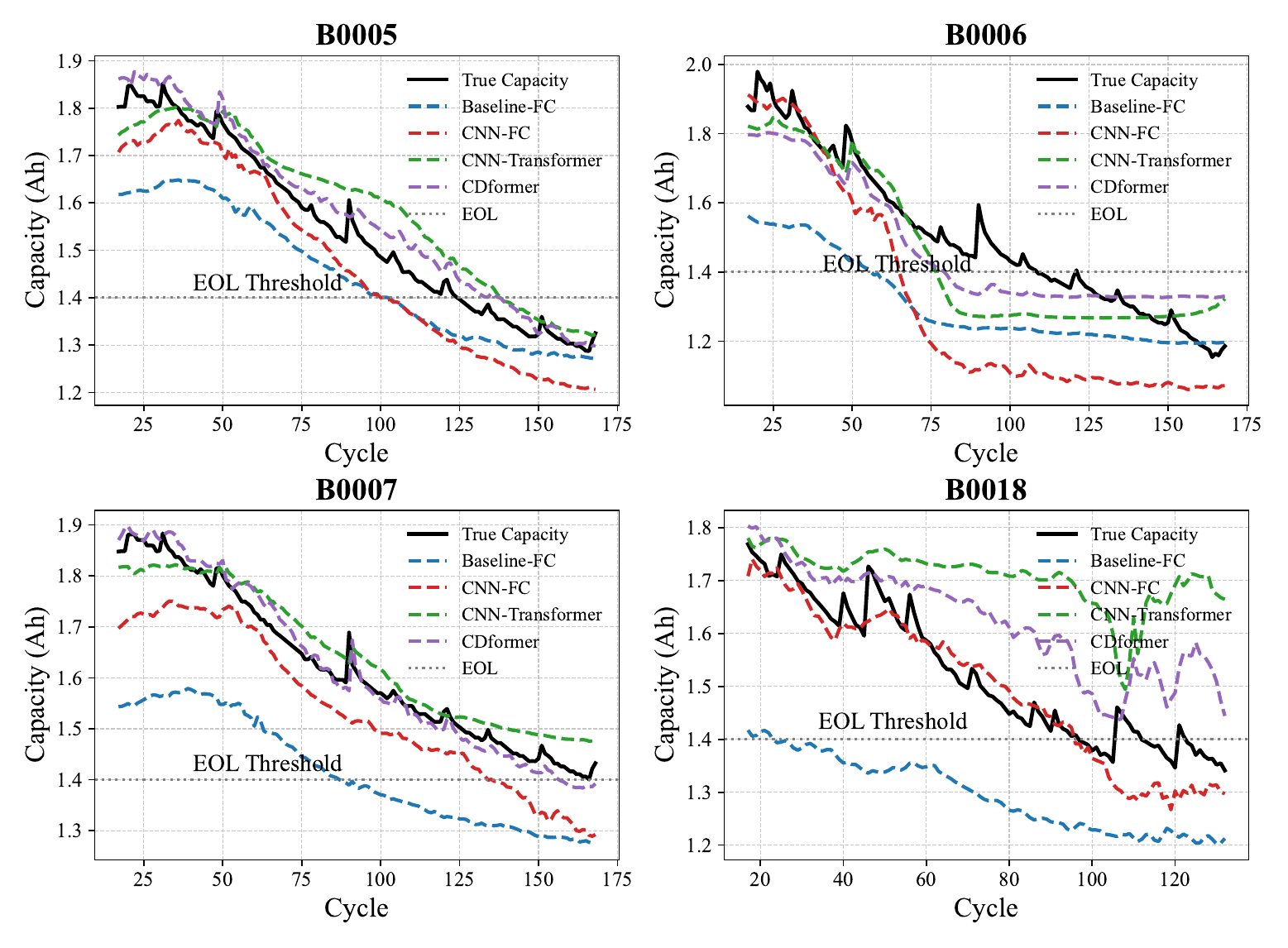}   % bottom: qualitative curves
    \caption{NASA dataset}
    \label{fig:variants_nasa_left}
  \end{subfigure}
  \hspace{0.01\linewidth}
  \begin{subfigure}{0.49\linewidth}
    \centering
    \includegraphics[width=\linewidth]{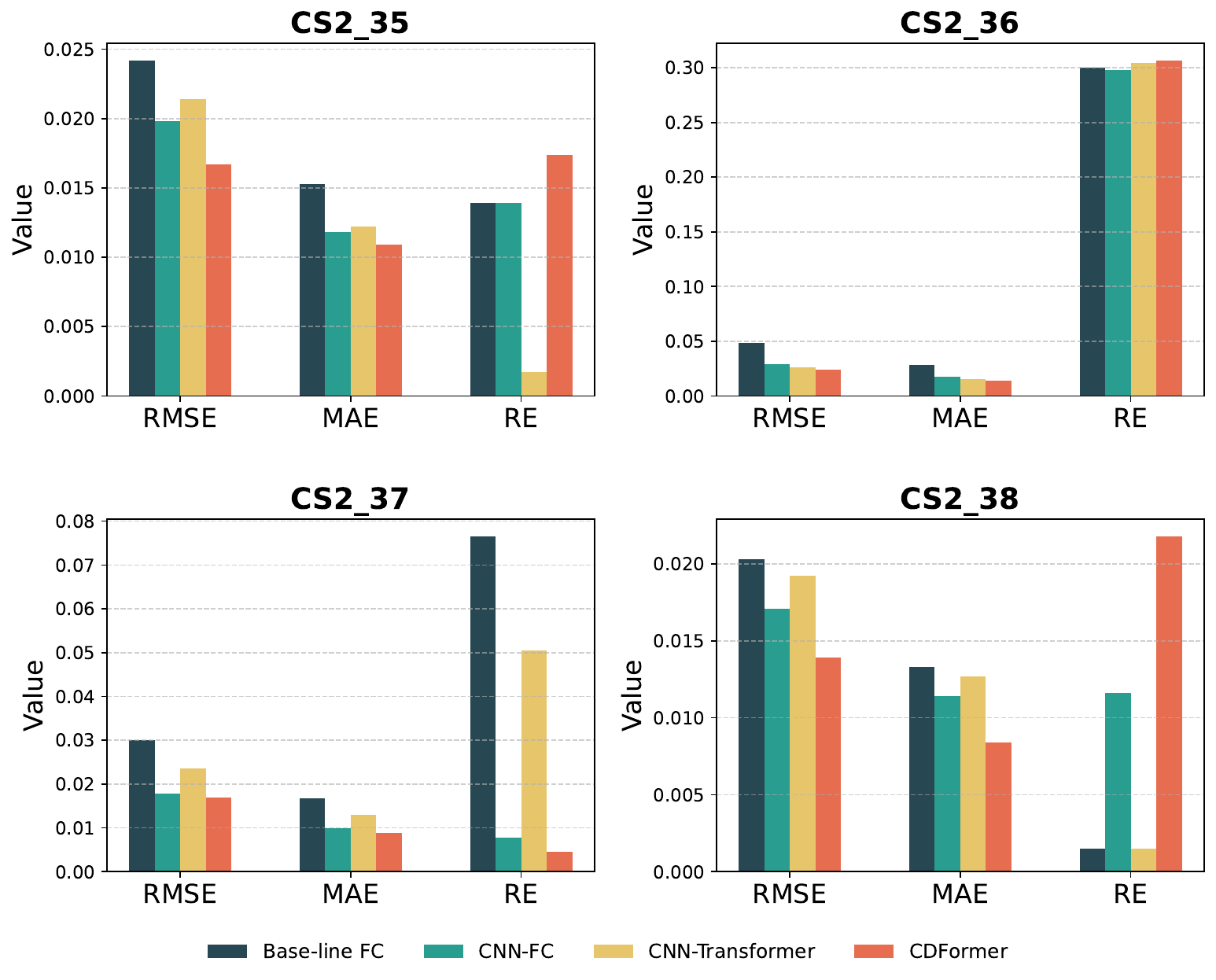} \\ % top: RMSE, MAE, RE
    \includegraphics[width=\linewidth]{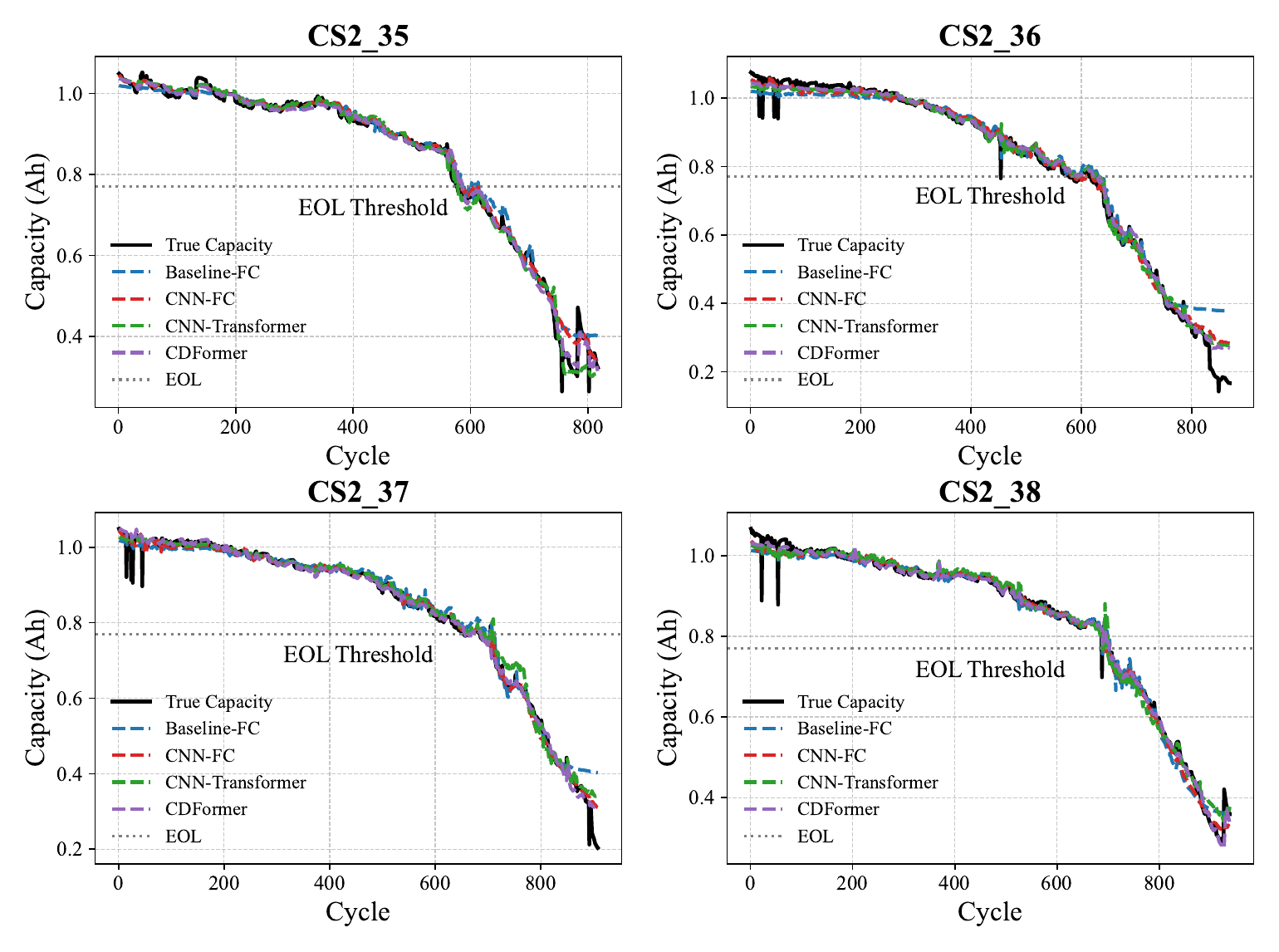}   % bottom: qualitative curves
    \caption{CALCE dataset}
    \label{fig:variants_calce_right}
  \end{subfigure}

  \caption{RMSE, MAE, and RE metrics (top) and predicted vs true capacity curves (bottom) of different deep learning-based variant models on each battery in the NASA and CALCE datasets.}
  \label{fig:variants_nasa_calce_combined}
\end{figure}
As shown in Fig.~\ref{fig:variants_nasa_calce_combined}, the upper subplots display RMSE, MAE, and RE metrics, and the lower subplots show predicted and true capacity curves for the NASA and CALCE datasets.

For the NASA dataset (Fig.~\ref{fig:variants_nasa_calce_combined}(a), left column), CDFormer achieved lower RMSE and MAE on most batteries. Improvements were particularly noticeable on B0005 and B0007. Although RE was slightly higher for some batteries, the absolute errors were smaller. In the capacity degradation curves, all models followed the general downward trend, while CDFormer aligned more closely with the ground truth across most cycles. It also tracked local fluctuations with smaller deviations.

For the CALCE dataset (Fig.~\ref{fig:variants_nasa_calce_combined}(b), right column), CDFormer obtained the lowest RMSE and MAE for all batteries, particularly on CS2\_35 and CS2\_38. It recorded the lowest RE on CS2\_37. The predicted curves captured rapid capacity variations more accurately than baseline models. The larger sample size and more complex degradation patterns in the CALCE dataset allowed CDFormer to learn richer representations of degradation behavior, supporting consistent predictions across batteries.

In summary, CDFormer achieved lower errors and closer curve alignment than baseline and variant models on both datasets. It modeled multi-scale temporal patterns effectively and tracked degradation processes accurately.

\subsection{Ablations}
\subsubsection{Effects of temporal data augmentation on CDFormer}
To systematically evaluate the impact of a composite temporal data augmentation strategy, comprising Gaussian noise, time warping, and time resampling, on the performance of CDFormer, we conducted a series of ablation experiments on both the NASA and CALCE datasets. The experiments involved five configurations: no augmentation, each individual augmentation method, and a combined augmentation strategy, aiming to quantify the contribution of each to the model's prediction accuracy.

\begin{table}[htbp]
\centering
\caption{\\Impact of temporal data augmentation on NASA and CALCE datasets.}
\label{tab:augment}
\begin{tabularx}{\linewidth}{l *{6}{>{\centering\arraybackslash}X}}
\toprule
\multirow{2}{*}{\textbf{temporal data augmentation}} &
\multicolumn{3}{c}{\textbf{NASA}} &
\multicolumn{3}{c}{\textbf{CALCE}} \\
\cmidrule(lr){2-4} \cmidrule(lr){5-7}
& \textbf{RMSE} $\downarrow$ & \textbf{MAE} $\downarrow$ & \textbf{RE} $\downarrow$ &
  \textbf{RMSE} $\downarrow$ & \textbf{MAE} $\downarrow$ & \textbf{RE} $\downarrow$ \\
\midrule
No Augmentation        & 0.0671 & 0.0586 & 0.2365 & 0.0179 & 0.0106 & 0.0877 \\
Gaussian Noise         & 0.0538 & 0.0447 & 0.1548 & 0.0166 & 0.0101 & 0.0869 \\
Time Warping           & 0.0585 & 0.0492 & 0.1162 & 0.0181 & 0.0111 & 0.0819 \\
Time Resampling        & 0.0666 & 0.0570 & 0.1184 & 0.0210 & 0.0125 & 0.0759 \\
All Combined (Ours)    & 0.0371 & 0.0301 & 0.1327 & 0.0169 & 0.0111 & 0.0817 \\
\bottomrule
\end{tabularx}
\end{table}
Table~\ref{tab:augment} shows that all augmentation techniques significantly enhanced performance on the relatively small and noisy NASA dataset. In particular, the combined augmentation setting achieved the best results, reducing RMSE, MAE, and RE by 44.7\%, 48.6\%, and 43.9\%, respectively, compared to the no-augmentation baseline. These results indicate that integrating multiple augmentation techniques can effectively increase data diversity and improve the model’s ability to capture temporal variations, thereby enhancing its prediction accuracy. On the CALCE dataset, which features a larger sample size and more complex degradation patterns, the performance gain from augmentation was less pronounced. Some single augmentation settings even led to slight increases in certain metrics. Although the combined augmentation still provided modest improvements, its marginal benefit was clearly reduced. This can be attributed to the rich feature distribution and sufficient training samples inherent in the CALCE dataset, which allow the model to achieve high prediction accuracy and generalization even without augmentation. Notably, in both datasets, the RE metric, reflecting the model’s accuracy near the batteries’ EOL, decreased significantly under most augmentation settings. This demonstrates that temporal data augmentation is particularly beneficial for improving prediction accuracy in the critical final stages of battery degradation.

\begin{figure}[htb]
  \centering
  \begin{minipage}[t]{0.49\linewidth}
    \centering
    \includegraphics[width=\linewidth]{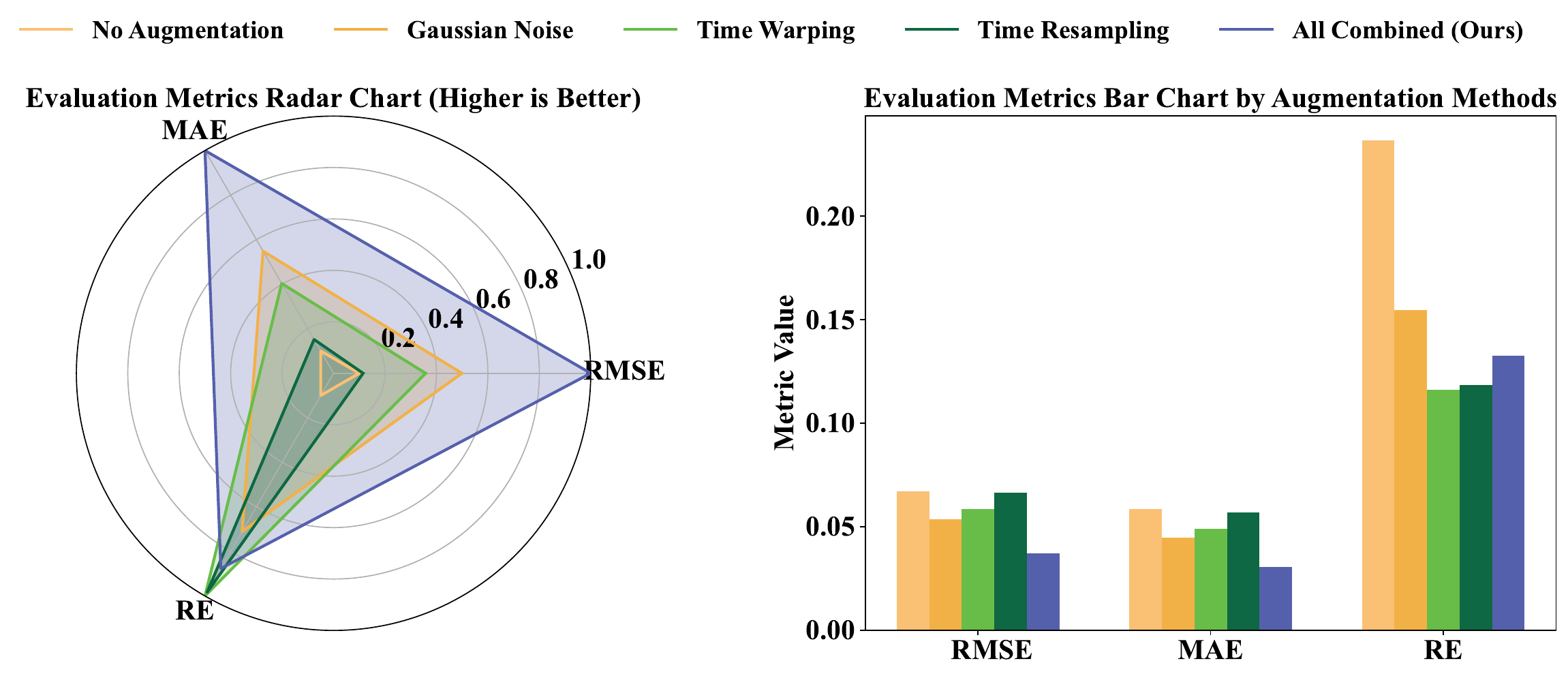}  % radar + bar NASA
    \includegraphics[width=\linewidth]{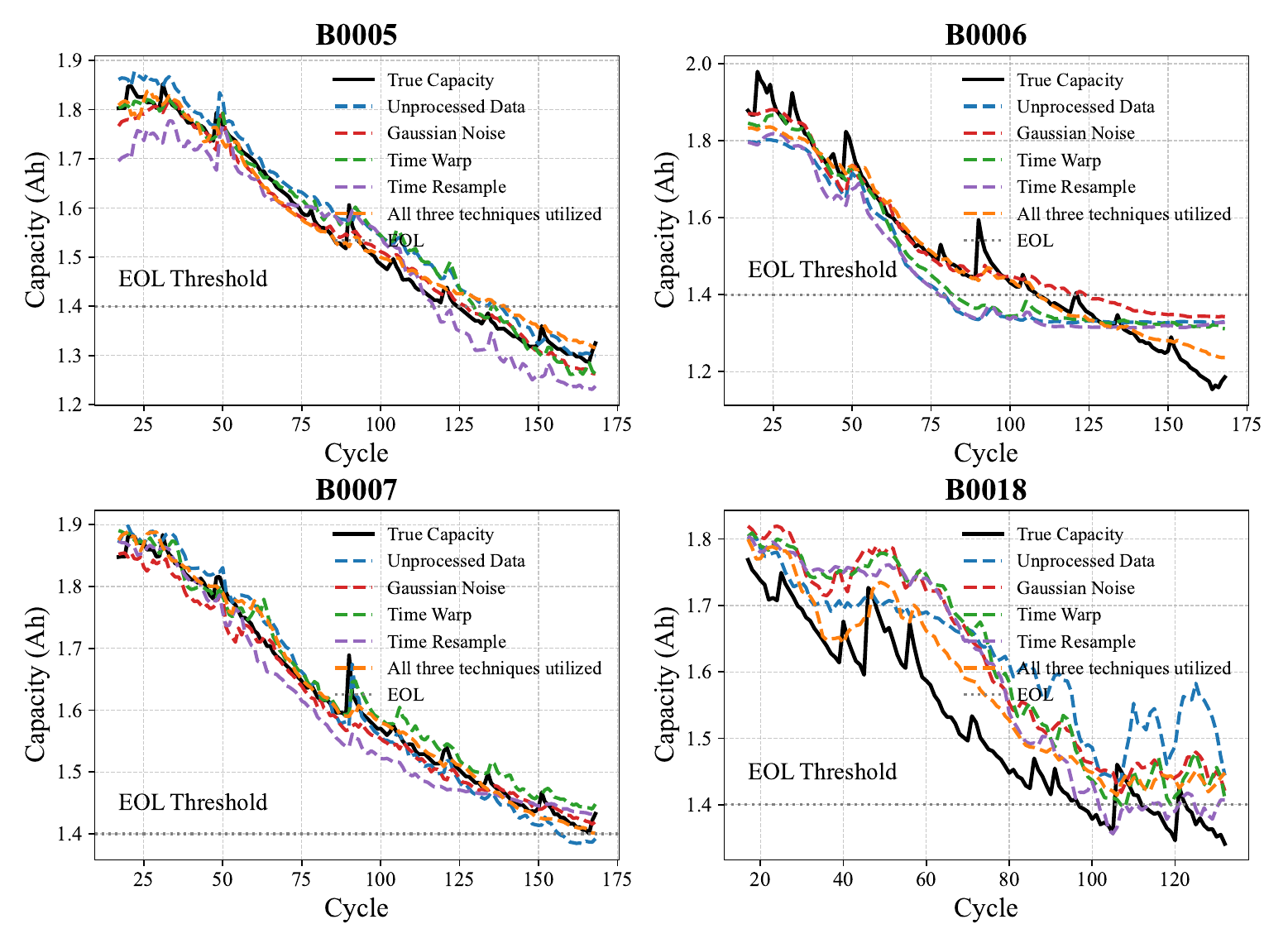} % predicted curves NASA
    \caption*{(a) NASA dataset}
    \label{fig:aug_nasa_combined}
  \end{minipage}
 \hspace{0.01\linewidth}
  \begin{minipage}[t]{0.49\linewidth}
    \centering
    \includegraphics[width=\linewidth]{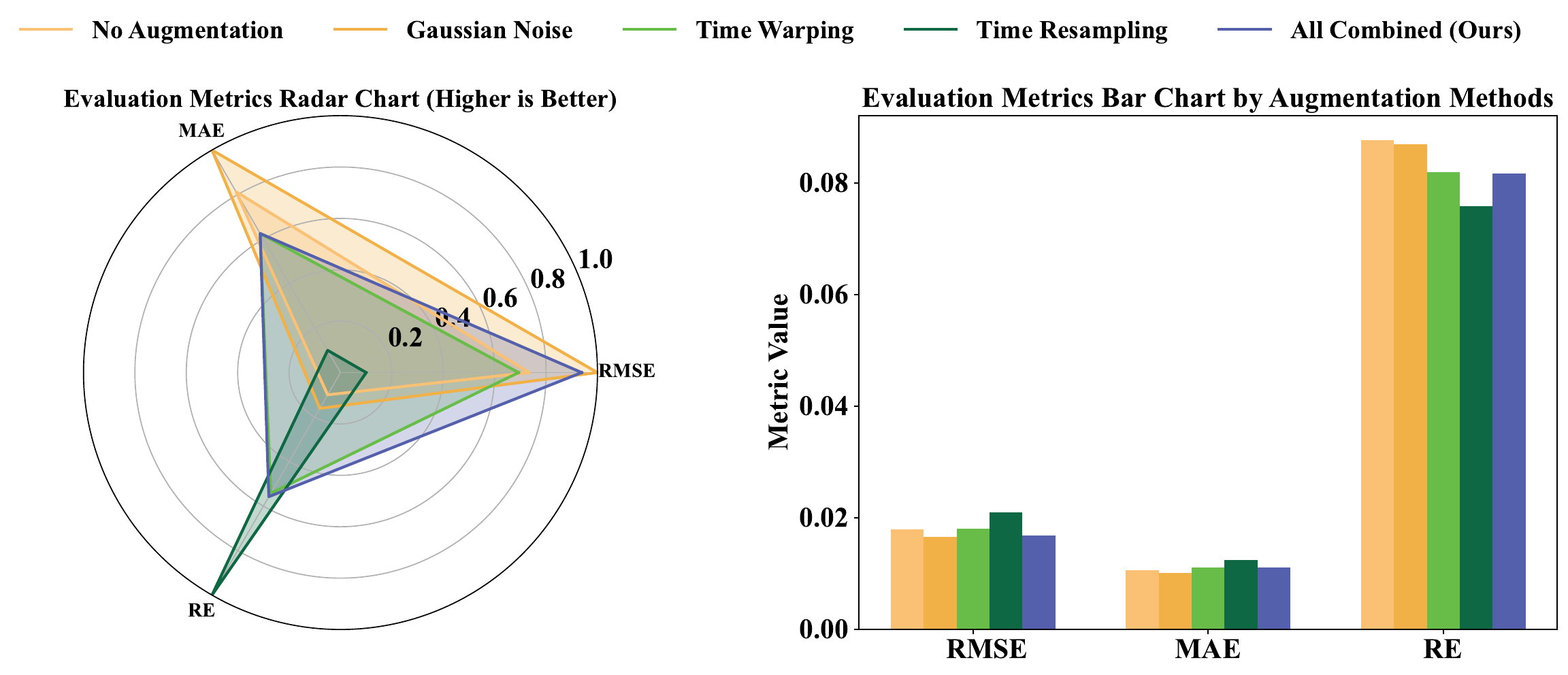}  % radar + bar CALCE
    \includegraphics[width=\linewidth]{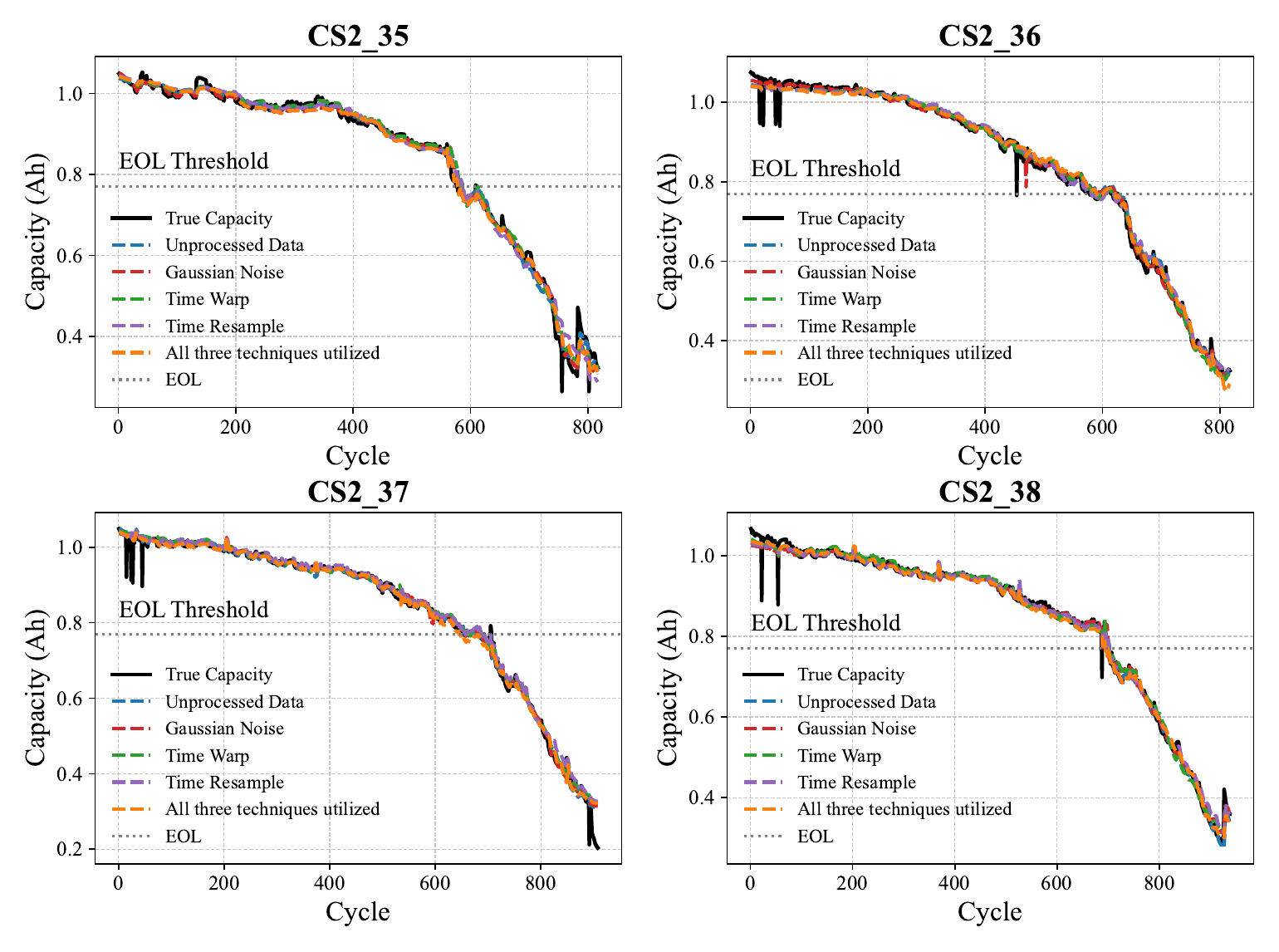} % predicted curves CALCE
    \caption*{(b) CALCE dataset}
    \label{fig:aug_calce_combined}
  \end{minipage}

  \caption{Visualization of temporal data augmentation results on the NASA and CALCE datasets. Top: normalized and inverted radar and bar charts of RMSE, MAE, and RE metrics; Bottom: predicted versus true capacity curves of CDFormer under different augmentation settings. Left column: NASA dataset; Right column: CALCE dataset.}
  \label{fig:data_augmentation_combined}
\end{figure}
To analyze the effects of temporal data augmentation, we visualized the performance of each augmentation method on the NASA and CALCE datasets using radar and bar charts, along with predicted capacity curves as shown in Figure~\ref{fig:data_augmentation_combined}. The upper subplots show the radar and bar charts, while the lower subplots show predicted versus true capacity curves. The left column corresponds to NASA, and the right column to CALCE.

On the NASA dataset, the combined augmentation strategy yielded the lowest RMSE and MAE, and a clear reduction in RE. Gaussian noise and time warping also decreased errors relative to the no-augmentation baseline, whereas time resampling exhibited less pronounced effects. In the predicted capacity curves, CDFormer with combined augmentation closely tracked the true capacity across most cycles, demonstrating smaller deviations than other configurations.

On the CALCE dataset, the differences among augmentation methods were less pronounced due to the larger sample size and lower noise levels. Time warping slightly reduced RE, while time resampling slightly increased errors. The combined augmentation still produced favorable results, although its advantage was less pronounced than on the NASA dataset. The predicted curves further demonstrated that CDFormer maintained consistent predictions across most cycles, in agreement with the quantitative results.

Overall, temporal data augmentation enhances prediction accuracy and robustness, particularly for datasets with smaller sample sizes or higher sensor noise, demonstrating its practical value in real-world battery monitoring scenarios.

\subsubsection{Effects of temporal data augmentation Across Model Variants}
To further assess the generalizability and effectiveness of a composite temporal data augmentation strategy, comprising Gaussian noise, time warping, and time resampling, we applied it to three distinct model variants: Baseline-FC, CNN-FC, and CNN-Trans. Quantitative evaluations were conducted on both the NASA and CALCE datasets to analyze the impact of this strategy across different architectures.

\begin{table}[htbp]
\setlength{\tabcolsep}{6pt}
\centering
\caption{\\Performance of model variants before and after temporal data augmentation on the NASA dataset}
\label{tab:NASA dataset}
\begin{tabularx}{\linewidth}{@{}lXXXXXX@{}}
\toprule
\multirow{2}{*}{\textit{Models}} & \multicolumn{3}{c}{\textit{Before temporal data augmentation}} & \multicolumn{3}{c}{\textit{After temporal data augmentation}} \\ \cmidrule(l){2-4} \cmidrule(l){5-7} 
 & RMSE $\downarrow$ & MAE $\downarrow$ & RE $\downarrow$ & RMSE $\downarrow$ & MAE $\downarrow$ & RE $\downarrow$ \\ 
\midrule
Baseline-FC & 0.1952 & 0.1822 & 0.5572 & 0.1814 & 0.1677 & 0.4005\\
CNN-FC & 0.1117 & 0.0981 & 0.2322 & 0.0748 & 0.0677 & 0.1990 \\
CNN-Transformer & 0.1059 & 0.0904 & 0.2264 & 0.0725 & 0.0606 & 0.1670 \\
CDFormer & 0.0671 & 0.0586 & 0.2365 & 0.0371 & 0.0301 & 0.1327 \\
\bottomrule
\end{tabularx}
\end{table}

\begin{table}[htbp]
\setlength{\tabcolsep}{6pt}
\centering
\caption{\\Performance of model variants before and after temporal data augmentation on the CALCE dataset}
\label{tab:CALCE dataset}
\begin{tabularx}{\linewidth}{@{}lXXXXXX@{}}
\toprule
\multirow{2}{*}{\textit{Models}} & \multicolumn{3}{c}{\textit{Before temporal data augmentation}} & \multicolumn{3}{c}{\textit{After temporal data augmentation}} \\ \cmidrule(l){2-4} \cmidrule(l){5-7} 
 & RMSE $\downarrow$ & MAE $\downarrow$ & RE $\downarrow$ & RMSE $\downarrow$ & MAE $\downarrow$ & RE $\downarrow$ \\ 
\midrule
Baseline-FC &  0.0309 & 0.0184 & 0.0980 & 0.0312 & 0.0188 & 0.0870 \\
CNN-FC & 0.0209 & 0.0127 & 0.0828 & 0.0213 & 0.0126 & 0.0799 \\
CNN-Transformer & 0.0225 & 0.0133 & 0.0896 & 0.0220 & 0.0134 & 0.0724 \\
CDFormer & 0.0179 & 0.0106 & 0.0877 & 0.0169 & 0.0111 & 0.0817 \\
\bottomrule
\end{tabularx}
\end{table}

As shown in Table~\ref{tab:NASA dataset}, the three models demonstrated performance improvements in the NASA dataset when augmented with the proposed techniques. RMSE, MAE, and RE values consistently decreased, highlighting the positive effect of augmentation under small-scale and noise-prone conditions. The RE metric, in particular, exhibited substantial reductions across models, indicating enhanced accuracy in predicting the critical end-of-life phase of battery degradation. On the CALCE dataset (Table~\ref{tab:CALCE dataset}), a similar trend was observed. The RE values of all three models were reduced after augmentation, demonstrating the robustness of the strategy in complex degradation scenarios. However, for the Baseline-FC model, a slight increase in RMSE and MAE was observed. This is likely due to the model’s simplistic architecture, which lacks the capacity to fully utilize the increased data diversity, potentially introducing redundancy or over-fitting when exposed to augmented inputs.

\begin{figure}[htbp]
    \centering
    \begin{subfigure}[t]{0.48\linewidth}
        \centering
        \includegraphics[width=\linewidth]{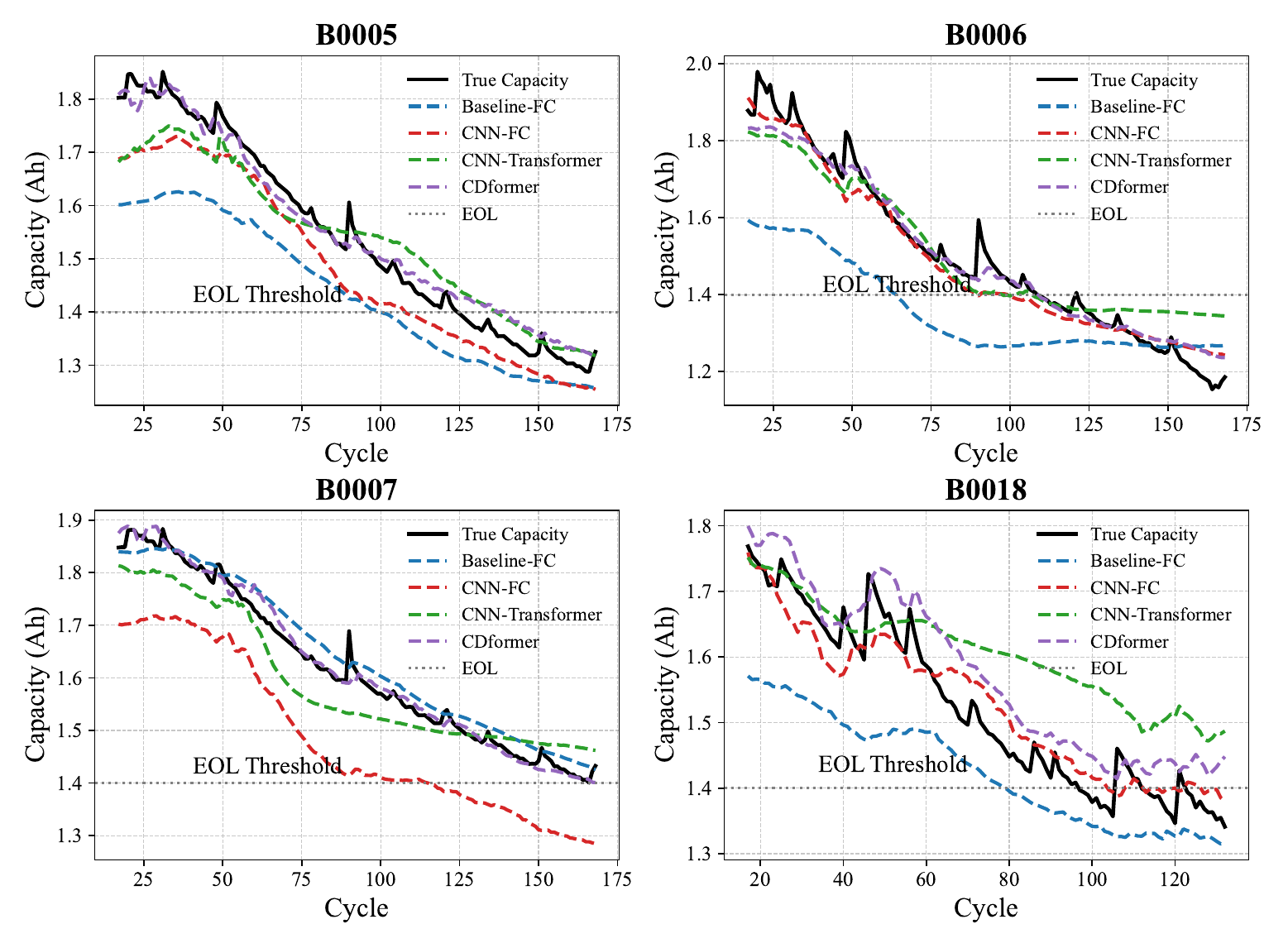}
        \caption{Impact of data augmentation on different model variants on the NASA dataset}
        \label{fig:variants_nasa}
    \end{subfigure}
    \hfill
    \begin{subfigure}[t]{0.48\linewidth} 
        \centering
        \includegraphics[width=\linewidth]{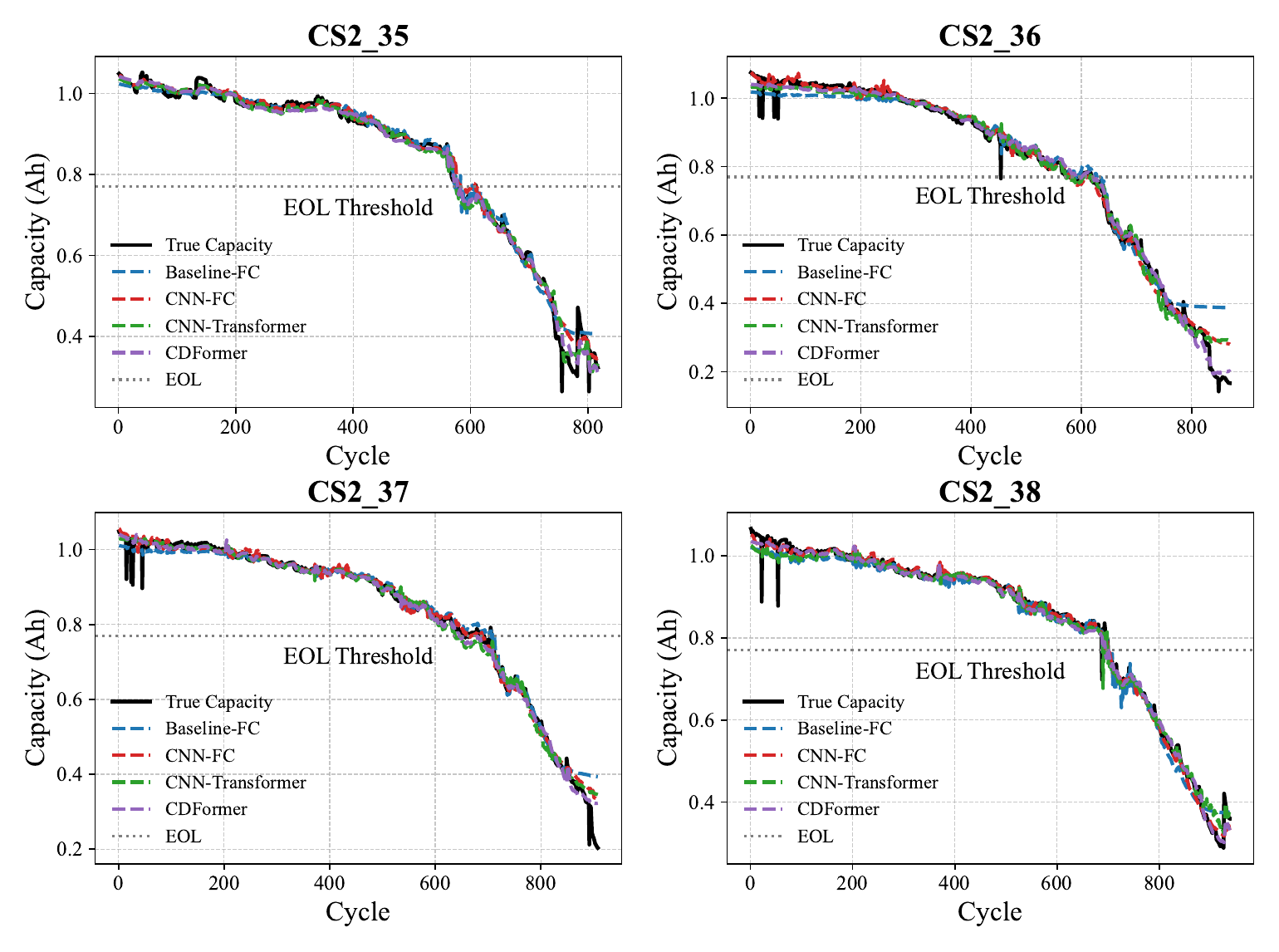}
        \caption{Impact of data augmentation on different model variants on the CALCE dataset}
        \label{fig:variants_calce}
    \end{subfigure}
    \caption{Comparison of the impact of data augmentation on different model variants for both NASA and CALCE datasets.}
    \label{fig:variants}
\end{figure}
To provide a more intuitive understanding, we visualized the predicted capacity curves of each model under the combined augmentation setting for both datasets, as shown in Figure~\ref{fig:variants}(a) and~\ref{fig:variants}(b). The curves clearly show that augmentation helps models generate smoother, more accurate predictions that closely align with the ground-truth capacity trajectories. Notably, CDFormer displayed the highest degree of consistency with the actual curves, particularly in complex or rapidly changing capacity regions. In conclusion, the proposed augmentation strategies not only significantly enhance CDFormer’s performance but also yield substantial improvements in other model variants. These findings validate the importance of data diversity in improving robustness and predictive accuracy, particularly in modeling long-term battery degradation dynamics and end-of-life behavior.

\subsubsection{Generalization under Diverse Temporal Feature Conditions}
To further investigate the model's ability to adapt to feature diversity, we applied the proposed temporal augmentation techniques to both CDFormer and several simplified model variants. As shown in Table~\ref{tab:nasa_data_augmentation} and Table~\ref{tab:calce_data_augmentation}, CDFormer consistently benefits from the enhanced data, achieving improved performance across all metrics. In contrast, several baseline models experienced degraded performance after applying the same augmentation techniques. This discrepancy highlights the importance of architectural design in effectively utilizing diverse temporal features. The superior performance of CDFormer under augmented inputs demonstrates its robustness and adaptability, while the weaker results of simpler models reflect their limited generalization capacity when exposed to increased input variability. These findings collectively underscore the necessity of robust temporal modeling mechanisms in battery RUL prediction tasks, particularly under complex and noisy real-world scenarios.
\begin{table}[ht]
\setlength{\tabcolsep}{6pt}
\centering
\caption{\\Impact of temporal data augmentation on single temporal feature models on the NASA dataset.}
\label{tab:nasa_data_augmentation}
\begin{tabularx}{\linewidth}{@{}lXXXXXX@{}}
\toprule
\multirow{2}{*}{\textit{Models}} & \multicolumn{3}{c}{\textit{Before temporal data augmentation}} & \multicolumn{3}{c}{\textit{After temporal data augmentation}} \\ \cmidrule(l){2-4} \cmidrule(l){5-7} 
 & RMSE $\downarrow$ & MAE $\downarrow$ & RE $\downarrow$ & RMSE $\downarrow$ & MAE $\downarrow$ & RE $\downarrow$ \\ 
\midrule
RNN~\cite{Catelani2021} & 0.1435 & 0.1205 & 0.3692 & 0.1429 & 0.1205 & 0.2880 \\
LSTM~\cite{ZhaoLSTM2022} & 0.1573 & 0.1356 & 0.3692 & 0.1574 & 0.1357 & 0.3692 \\
GRU~\cite{Lin2023} & 0.0792 & 0.0675 & 0.3002 & 0.0793 & 0.0675 & 0.3002 \\
DeTransformer~\cite{Chen2022} & 0.0863 & 0.0746 & 0.2609 & 0.0865 & 0.0747 & 0.2586 \\
SA-LSTM~\cite{WangLSTM2023} & 0.0897 & 0.0751 & 0.2013 & 0.0898 & 0.0753 & 0.1990 \\
AttMoE~\cite{ChenAttMoE2024} & 0.0847 & 0.0730 & 0.2849 & 0.0851 & 0.0731 & 0.2845 \\
CDFormer & 0.0671 & 0.0586 & 0.2365 & 0.0371 & 0.0301 & 0.1327 \\
\bottomrule
\end{tabularx}
\end{table}

\begin{table}[ht]
\setlength{\tabcolsep}{6pt}
\centering
\caption{\\Impact of temporal data augmentation on single temporal feature models on the CALCE dataset.}
\label{tab:calce_data_augmentation}
\begin{tabularx}{\linewidth}{@{}lXXXXXX@{}}
\toprule
\multirow{2}{*}{\textit{Models}} & \multicolumn{3}{c}{\textit{Before temporal data augmentation}} & \multicolumn{3}{c}{\textit{After temporal data augmentation}} \\ \cmidrule(l){2-4} \cmidrule(l){5-7} 
 & RMSE $\downarrow$ & MAE $\downarrow$ & RE $\downarrow$ & RMSE $\downarrow$ & MAE $\downarrow$ & RE $\downarrow$ \\ 
\midrule
RNN~\cite{Catelani2021} & 0.0534 & 0.0423 & 0.1418 & 0.0535 & 0.0423 & 0.1418 \\
LSTM~\cite{ZhaoLSTM2022} & 0.0720 & 0.0487 & 0.1466 & 0.072 & 0.0487 & 0.1470 \\
GRU~\cite{Lin2023} & 0.0973 & 0.0756 & 0.1646 & 0.0975 & 0.0757 & 0.1646 \\
DeTransformer~\cite{Chen2022} & 0.0616 & 0.0481 & 0.1282 & 0.0619 & 0.0484 & 0.1271 \\
SA-LSTM~\cite{WangLSTM2023} & 0.0324 & 0.0263 & 0.1097 & 0.0323 & 0.0262 & 0.1103 \\
AttMoE~\cite{ChenAttMoE2024} & 0.0250 & 0.0180 & 0.1345 & 0.0253 & 0.0183 & 0.1349 \\
CDFormer & 0.0179 & 0.0106 & 0.0877 & 0.0169 & 0.0111 & 0.0817 \\
\bottomrule
\end{tabularx}
\end{table}

\section{Conclusion}
In this study, we proposed CDFormer, a hybrid deep learning framework. It integrates CNNs, DRSNs, and Transformer encoders to predict lithium-ion battery RUL. CDFormer applies Gaussian noise and two novel temporal data augmentation techniques. This design improves its generalization and robustness under complex operating conditions.

Extensive experiments on the NASA and CALCE datasets show that CDFormer outperforms several baseline and state-of-the-art models. Ablation studies indicate that the proposed augmentation techniques contribute to better predictive performance.

Future work will focus on incorporating multi-modal sensor data and exploring the integration of CDFormer with physics-based models or digital twins to achieve joint estimation of SOH and RUL with improved interpretability.

Overall, the findings highlight CDFormer as an effective and extensible framework for battery health management, supporting predictive maintenance and contributing to improved safety and efficiency in energy storage systems.

\bibliographystyle{elsarticle-num}
\bibliography{references}

\end{document}